\documentclass[journal]{IEEEtran}
\usepackage{abbr}
\usepackage{times}
\usepackage{epsfig}
\usepackage{graphicx}
\usepackage{amsmath}
\usepackage{amssymb}
\usepackage[norelsize, linesnumbered, ruled, lined, boxed, commentsnumbered]{algorithm2e}
\usepackage[table,xcdraw]{xcolor}
\usepackage{multirow}
\usepackage{adjustbox}
\usepackage{tabularx}
\usepackage{booktabs}
\usepackage{array}
\usepackage{url}

\newcommand{\aq}[1]{{#1}}
\newcommand{\wty}[1]{{#1}}

\ifCLASSINFOpdf
\else
\fi
\begin{document}
\title{\wty{E2Style: Improve the Efficiency and Effectiveness of StyleGAN Inversion}}

\author{Tianyi~Wei$^*$,~Dongdong~Chen$^*$,~Wenbo~Zhou$^\dagger$,~Jing~Liao,~Weiming~Zhang,~Lu~Yuan,~Gang~Hua, ~\IEEEmembership{Fellow,~IEEE}, Nenghai~Yu
        
\IEEEcompsocitemizethanks{
\IEEEcompsocthanksitem Tianyi Wei, Wenbo Zhou, Weiming Zhang, Nenghai Yu are with University of Science and Technology of China, Hefei, Anhui 230026, China. E-mail: \{bestwty@mail., welbeckz@, zhangwm@, ynh@\}ustc.edu.cn
\IEEEcompsocthanksitem  Dongdong Chen and Lu Yuan are with Microsoft Cloud AI, Redmond, Washington 98052, USA. Email: cddlyf@gmail.com, luyuan@microsoft.com
\IEEEcompsocthanksitem Jing Liao is with the Department of Computer Science, City University of Hong Kong. Email: jingliao@cityu.edu.hk
\IEEEcompsocthanksitem Gang Hua is with Wormpex AI Research LLC, WA 98004, US. E-mail: ganghua@gmail.com
\IEEEcompsocthanksitem $^*$ Tianyi Wei and Dongdong Chen are co-first authors, $\dagger$ Wenbo Zhou is the corresponding author.
}}

% \markboth{Journal of \LaTeX\ Class Files,~Vol.~14, No.~8, August~2015}%
% {Shell \MakeLowercase{\textit{et al.}}: Bare Demo of IEEEtran.cls for IEEE Journals}



\maketitle

\begin{abstract}
This paper studies the problem of StyleGAN inversion, which plays an essential role in enabling the pretrained StyleGAN to be used for real image editing tasks. The goal of StyleGAN inversion is to find the exact latent code of the given image in the latent space of StyleGAN. This problem has a high demand for quality and efficiency.
Existing optimization-based methods can produce high-quality results, but the optimization often takes a long time. On the contrary, forward-based methods are usually faster but the quality of their results is inferior. In this paper, we present a new feed-forward network ``E2Style" for StyleGAN inversion, with significant improvement in terms of efficiency and \wty{effectiveness}. In our inversion network, we introduce: 1) a shallower backbone with multiple efficient heads across scales; 2) multi-layer identity loss and multi-layer face parsing loss to the loss function; and 3) multi-stage refinement. Combining these designs together forms an \wty{effective} and efficient method that exploits all benefits of optimization-based and forward-based methods. Quantitative and qualitative results show that our E2Style performs better than existing forward-based methods and comparably to state-of-the-art optimization-based methods while maintaining the high efficiency as well as forward-based methods. Moreover, a number of real image editing applications demonstrate the efficacy of our E2Style. Our code is available at \url{https://github.com/wty-ustc/e2style}%on our project page:.} 
%~\url{https://wty-ustc.github.io/inversion}
\end{abstract}

\begin{IEEEkeywords}
StyleGAN inversion, Effectiveness, Efficiency
\end{IEEEkeywords}

% \fi
\IEEEpeerreviewmaketitle

\section{Introduction}

\IEEEPARstart{G}{AN} inversion aims to invert a real image back into the latent space of a pretrained GAN model, such as StyleGAN~\cite{Karras2019ASG,Karras2020AnalyzingAI}, for the image to be faithfully reconstructed from the inverted code by the generator. It not only provides an alternative flexible image editing framework but also helps reveal the mechanism underneath deep generative models. As an emerging technique to bridge the pretrained GAN model and real image editing tasks \cite{Shen2020InterpretingTL,Abdal2019Image2StyleGANHT,Abdal2020Image2StyleGANHT}, GAN inversion has a high demand for quality and efficiency.

Recently, numerous GAN inversion methods~\cite{Abdal2019Image2StyleGANHT,Abdal2020Image2StyleGANHT,Nitzan2020FaceID,Guan2020CollaborativeLF,Yang2021ADN} have been proposed and shown strong competence in performing meaningful manipulations of human faces in the latent space. They can be mainly categorized as optimization-based \cite{Abdal2019Image2StyleGANHT,Abdal2020Image2StyleGANHT,Pan2020ExploitingDG} and forward-based \cite{Richardson2020EncodingIS,Nitzan2020FaceID,Guan2020CollaborativeLF}. The optimization-based approach optimizes the latent code directly for a given single image by back-propagation. It is capable of producing high-quality inversion but the optimization process is too time-consuming, thus greatly limiting its real-time applications. The forward-based method uses an encoder network to learn the mapping from the image space to the latent space, where only one feed-forward pass is required in the inference, providing higher efficiency for real-time applications. However, it suffers from lower reconstruction quality, and its network structure is usually large and complex. Besides, the hybrid approach~\cite{Zhu2020InDomainGI,Bau2019SeeingWA,Bau2019SemanticPM,Yang2021ADN} that incorporates the optimization on top of the forward network mitigates the problem of quality, but the time cost is greatly increasing.

\begin{figure*}[t]
	\centering
	\includegraphics[width=\textwidth]{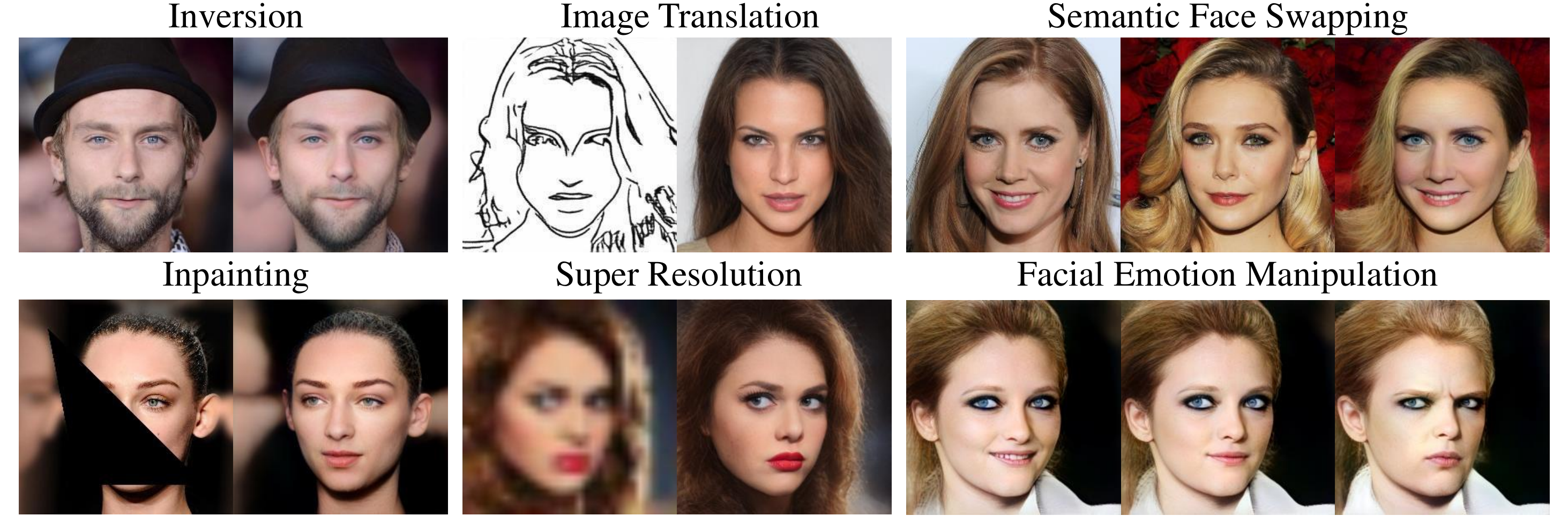} 
	\caption{E2Style can be applied to a large number of real image editing tasks, such as image restoration, image manipulation, image translation, \etc.} 
	\label{fig:teaser}
\end{figure*}

In this paper, we focus on StyleGAN inversion and propose a new feed-forward network \textbf{``E2Style"}, which significantly improves existing approaches in terms of both \textbf{E}fficiency and \textbf{E}ffectiveness. 
%Such a baseline is quite simple but surprisingly effective. 

For the network structure design, we have two insights. First, it is critical to explicitly decouple the information of different layers. In StyleGAN, the latent code of different layers corresponds to different semantic levels, and the later part of the latent code corresponds to the lower level feature information. Existing frameworks such as IDGI \cite{Zhu2020InDomainGI} usually only use the final feature compressed by the encoder to regress the latent code, which forces various semantic information to couple together in the final feature. Unlike them, our encoder network considers a {\em hierarchical} structure for the latent code prediction. In this way, feature vectors extracted from various spatial levels of the encoder can correspond to different semantic levels of details from the pretrained StyleGAN generator, achieving semantic correspondence and reducing the learning difficulty. Meanwhile, we find it is not that the deeper the encoder, the better the inversion will be. And a {\em shallower} encoder is sufficient for the latent code prediction.

Second, the regression should be performed with as little information loss as possible. The pSp \cite{Richardson2020EncodingIS} downsamples the features from the encoder through several convolutional layers until the resolution is $ 1\times1 $, and then regresses the latent code through the full-connected layer, which causes a large degree of information loss. Instead, we adopt a shared {\em efficient prediction head} at each level, which only consists of a global average pooling layer with varied sizes to levels and a full-connected layer. For features corresponding to the low-level information we use a larger size, which preserves more information and therefore brings performance gains. This efficient prediction head is more lightweight and efficient than the complex and independent heads used in pSp \cite{Richardson2020EncodingIS} that consists of a series of convolutional layers. With these two  network architecture improvements, our network becomes smaller but outperforms IDGI \cite{Zhu2020InDomainGI} and pSp \cite{Richardson2020EncodingIS}.

%Specifically, we introduce two core sections to the network architecture. First, our encoder network considers a {\em hierarchical} structure for the latent code prediction, rather than the solely last layer of the encoder for the prediction as performed in existing forward networks~\cite{Zhu2020InDomainGI}. In this way, feature vectors extracted from various spatial levels of the encoder can correspond to different semantic levels of details from the pretrained StyleGAN generator. Meanwhile, we find it is not that the deeper the encoder, the better the inversion will be. And a {\em shallower} encoder is sufficient for the latent code prediction. Second, we adopt a shared {\em efficient prediction head} at each level, which only consists of a global average pooling layer with varied sizes to levels and a full-connected layer. It is more lightweight and efficient than the complex and independent heads used in \cite{Richardson2020EncodingIS} that consists of a series of convolutional layers. 

Besides the architecture, we introduce two new losses into the objective function for better effectiveness. One is {\em multi-layer identity loss}, which provides stronger semantic alignment supervision compared to single-layer identity loss used in pSp~\cite{Richardson2020EncodingIS} and thus greatly improves the identity consistency between the reconstructed image and the input real image. The other is {\em multi-layer face parsing loss}, which helps capture local facial details (\eg, eyes, mouth) for a more fine-grained reconstruction.

% \wty{As some works have claimed \cite{Richardson2020EncodingIS, Bayat2020InverseMO}, it is very challenging to achieve the one-step prediction of latent code using encoders. }
In order to further reduce the quality gap between the single stage prediction of latent code and the ideal prediction, we propose a {\em multi-stage refinement} learning approach to progressively predict the residual of the latent code through multiple passes of the encoders to achieve better inversion quality. Specifically, we introduce a recursive forward refinement phase in our framework, which learns the residual of the latent code output from the first phase to find a more optimal solution that exists around that latent code. E2Style shares the similar spirit to IDGI's prediction-before-optimization method \cite{Zhu2020InDomainGI}, but we improve the initial inverted latent code by introducing a refinement network. This improves the performance without a significant increase in time consumption.

\begin{figure*}[t]
	\centering
	\includegraphics[width=\textwidth]{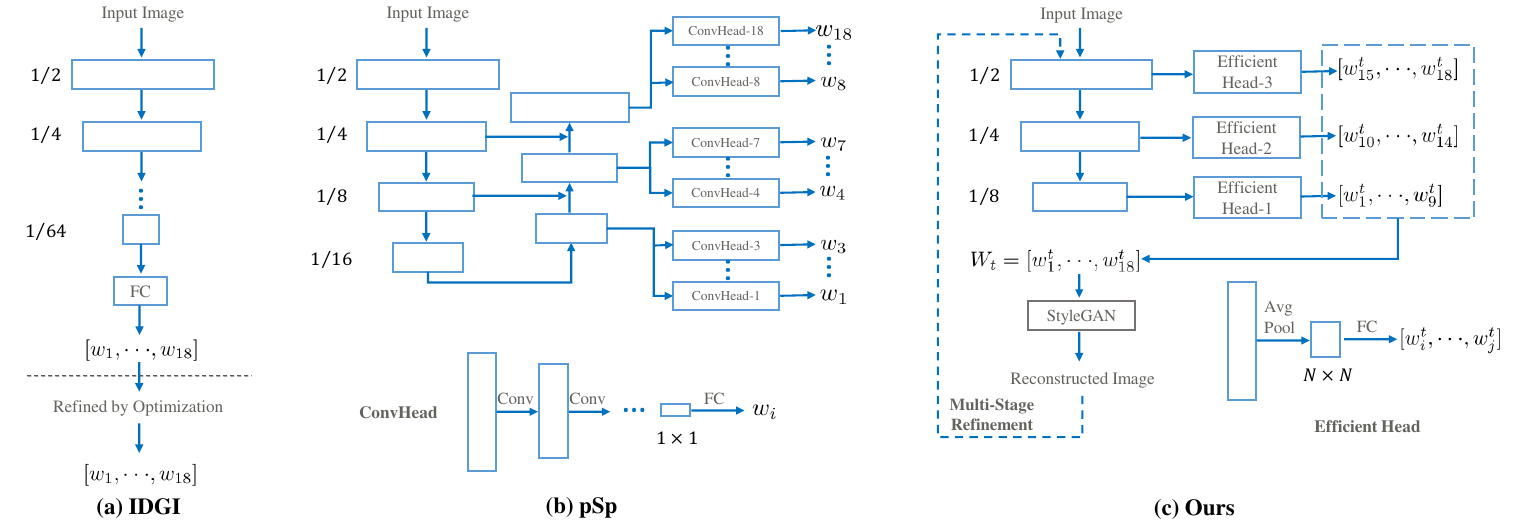} 
	\caption{Network structure comparison with other methods. IDGI \cite{Zhu2020InDomainGI} uses a deep backbone network and pSp \cite{Richardson2020EncodingIS} employs $18$ high-overhead ConvHeads to predict the single layer latent code separately. Compared with them, we design a shallower backbone and a more efficient prediction head. We also introduce the purely feed-forward-based multi-stage refinement, which improves the performance without bringing a significant increase in time consumption.} 
	\label{fig:modelcompare}
\end{figure*}

Qualitative and quantitative experiments show that E2Style substantially outperforms existing forward-based methods and even achieves performance comparable to the state-of-the-art optimization-based methods. A number of applications including secure deep hiding, image manipulation, image restoration, and image translation evidence the generalization of E2Style for real-time image editing tasks, illustrated in Figure \ref{fig:teaser}.

To summarize, our contributions are summarized as follows:
	\begin{itemize}
		\item We design an efficient network structure for StyleGAN inversion, which has fewer network parameters and better performance than existing forward-based methods.
		
		\item To further improve the inversion effectiveness, we utilize multi-layer identity loss and multi-layer face parsing loss as well as introduce a recursive forward refinement phase.
		
		\item We propose a new application of StyleGAN inversion: secure deep hiding. And various real image editing applications demonstrate the power of our approach.
		
\end{itemize}

\section{Related Work}
\subsection{Generative Adversarial Networks (GANs).} Since GANs were first proposed by Goodfellow \etal \cite{Goodfellow2014GenerativeAN} in $ 2014 $, it has evolved considerably in terms of training strategies \cite{Karras2020AnalyzingAI,Tran2021OnDA}, loss functions \cite{Ansari2020ACF,Arjovsky2017WassersteinGA, mao2017least}, regularizations \cite{Miyato2018SpectralNF,Brock2019LargeSG, qin2020does}, and network structures \cite{Schnfeld2020AUB,Gulrajani2017ImprovedTO}. Today's popular GANs have demonstrated amazing capabilities in many computer vision tasks \cite{wan2020bringing,Tan2019ImprovedAF,Hsu2019SiGANSG,Xu2020MEFGANMI,Chen2019GatedGANAG,Gao2020RPDGANLT,Lucas2019GenerativeAN, liu2021fiss, zhang2020mu, li2021predicting}, especially image synthesis \cite{tan2020michigan,tan2021efficient,tan2021diverse}. BigGAN \cite{Brock2019LargeSG} can successfully generate high-fidelity, diverse samples for complex and large-scale datasets like ImageNet \cite{Deng2009ImageNetAL} by only feeding the class condition. ProGAN \cite{Karras2018ProgressiveGO} and StyleGAN \cite{Karras2019ASG,Karras2020AnalyzingAI} can synthesize high-resolution images up to $ 1024\times1024$ with a progressive upsample network. To control the GAN synthesis process, some works explore the semantic editing of latent codes by finding semantic directions on their latent space in a supervised \cite{Shen2020InterpretingTL} or unsupervised \cite{Shen2020ClosedFormFO, Hrknen2020GANSpaceDI} manner. 
Besides, built upon these pretrained GAN models, a lot of real image editing tasks can also be conducted in the latent space, such as super resolution, face attribute manipulation. To enable these tasks, GAN inversion acts as the intermediate bridge. Specifically, the target real image will be firstly inverted into the latent space via such inversion techniques, then be edited in the latent space. In this paper, we follow existing methods \cite{wang2021cross,Richardson2020EncodingIS,Abdal2019Image2StyleGANHT,Abdal2020Image2StyleGANHT,Guan2020CollaborativeLF,Zhu2020InDomainGI} and choose StyleGAN as the inversion target, but the proposed feed-forward GAN inversion network and the accompanying design principles can be easily adapted to other GAN models. 

\subsection{GAN Inversion.} Existing GAN inversion methods can be subdivided into optimization-based \cite{Abdal2019Image2StyleGANHT,Abdal2020Image2StyleGANHT,Gu2020ImagePU,Pan2020ExploitingDG}, forward-based \cite{Richardson2020EncodingIS,Nitzan2020FaceID,Bartz2020OneMT,Guan2020CollaborativeLF} and hybrid \cite{Zhu2020InDomainGI,Bau2019SeeingWA,Bau2019SemanticPM,Yang2021ADN, yang2021inversion} approaches.

 The optimization-based approach directly optimizes the latent code to minimize the reconstruction loss, so that the synthetic image generated by the inverted latent code is as similar as possible to the target real image. For example, Pan \etal \cite{Pan2020ExploitingDG} optimize the latent code while fine-tuning the parameters of the generator. Image2StyleGAN++ \cite{Abdal2020Image2StyleGANHT} takes alternative optimizations of StyleGAN's latent code and noise space to further improve the inversion performance. Gu \etal \cite{Gu2020ImagePU} argue that the mapping of a single image to the low-dimensional latent code is necessarily lossy and therefore uses multiple latent codes to reconstruct a single image, thus alleviating this situation. Although such optimization-based methods can achieve good reconstruction results, the optimization process is too time-consuming and requires even thousands of iterations and ten minutes for some examples, especially when the initialization point is too far from the ideal embedding, thus greatly limiting its real-time application. 
%Although optimization-based methods can achieve good reconstruction results, the long optimization time limits their application. 

To speed up, the forward-based methods directly use an encoder network to learn the mapping from image space to the latent space. For example, Nitzan \etal \cite{Nitzan2020FaceID} use two encoders to encode identity information and attribute information separately. Guan \etal \cite{Guan2020CollaborativeLF} propose a novel collaborative learning framework to train the encoder in an unsupervised manner. 
 Concurrent with our work, ReStyle \cite{alaluf2021restyle} extends the encoder-based inversion method by introducing an iterative refinement mechanism. This idea is similar to our multi-stage refinement, but we use a separate encoder at each stage and outperform ReStyle in terms of all evaluation metrics. The pSp \cite{Richardson2020EncodingIS} designs a feature pyramid network with independent inversion heads for each latent code and introduces the identity loss as an additional constraint.
%  \wty{However, most of the current learning-based methods can only produce reasonable reconstructed images for synthetic images sampled from latent space or images with similar distribution to the training dataset, and its performance is often unsatisfactory once the input is the real-world image.}
  Different from these methods, we introduce a simple feed-forward network with improved efficiency and quality. 
%  \wty{In addition, we can achieve accurate reconstruction of real-world images.}
 Moreover, the newly proposed multi-layer identity loss, local parsing loss, and multi-stage refinement should also generalize to these methods.

Furthermore, the hybrid approach combines forward-based and optimization-based methods. Zhu \etal \cite{Zhu2020InDomainGI} use an encoder to generate an initial latent code, and the following optimizer uses it as the starting point to further refine the latent code. \wty{Yang \etal \cite{yang2021inversion} follow the prediction-then-optimization approach of \cite{Zhu2020InDomainGI} except that a detached dual-channel domain encoder is proposed.} Similarly, DNI \cite{Yang2021ADN} uses pSp \cite{Richardson2020EncodingIS} as the domain-guided encoder to predict the initial latent code, and then a noise optimization mechanism is implemented to capture high-frequency details and further improve the reconstruction quality.
Our multi-stage refinement is inspired by these methods but improves the inversion quality with purely feed-forward-based refinements, thus our speed is much faster than such hybrid approaches.

\section{Proposed Method}

In this section, we will first briefly introduce some representative feed-forward networks designed for GAN inversion. Then we will elaborate on the details of our \textbf{``E2Style"} for better efficiency and effectiveness from three aspects: network design, improved loss functions, and multi-stage refinement.

\begin{figure*}[t]
	\centering
	\includegraphics[width=\textwidth]{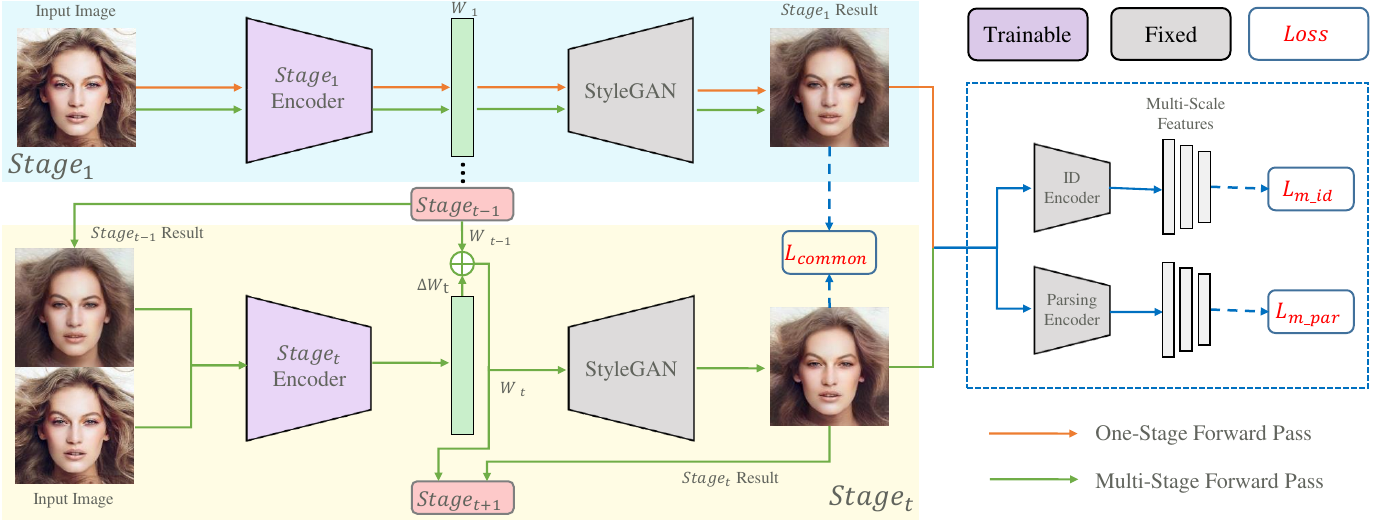} 
	\caption{The overview of our feed-forward network \textbf{E2Style} for StyleGAN inversion. Data flow is marked by solid lines and losses by dashed ones. Only the encoder is trainable. Except for the first stage encoder, all other encoders take the original image and the output image of the previous stage as input and output the residual of the latent code that was predicted in the previous stage.} 
	\label{fig:modelandlossfigure}
\end{figure*}

%\subsection{Background}
%
%\wty{
%	In recent years, Generative Adversarial Networks (GANs) have been developed rapidly, typically represented by StyleGAN \cite{Karras2019ASG,Karras2020AnalyzingAI}, which can synthesize high-resolution, high-fidelity realistic images. StyleGAN employs a Multi-Layer Perceptron (MLP) mapping network to map random noise inputs into $ \mathcal{W} $ latent space. Several studies \cite{Collins2020EditingIS, Shen2020InterpretingTL, Goetschalckx2019GANalyzeTV, Jahanian2020OnT} have demonstrated that StyleGAN spontaneously learns to encode rich semantics within $ \mathcal{W} $ space during training, and thus $ \mathcal{W} $ exhibits good semantic decoupling properties. However, training such a GAN requires unaffordable time, computational resources and dataset size for the normal user.}
%
%\wty{
%	Considering the consensus \cite{Abdal2019Image2StyleGANHT,Abdal2020Image2StyleGANHT,Guan2020CollaborativeLF,Richardson2020EncodingIS} from previous work, we choose $ \mathcal{W+} $ as the target latent space, which is defined by a cascade of $ 18 $ different $ 512 $-dimensional vectors $ w \in \mathcal{W} $.}

\subsection{Network Structure Design}
In Figure \ref{fig:modelcompare}, we show two representative network structures designed in IDGI \cite{Zhu2020InDomainGI} and pSp \cite{Richardson2020EncodingIS} for StyleGAN inversion. All these methods choose the $ \mathcal{W+} $ space \cite{Abdal2019Image2StyleGANHT, Abdal2020Image2StyleGANHT, Richardson2020EncodingIS, Guan2020CollaborativeLF, Zhu2020InDomainGI} as the target latent space to invert, which is demonstrated to be more suitable for GAN inversion than the original $\mathcal{W}$ space. It is defined by the cascade of $ 18 $ different $ 512 $-dimensional vectors $ [w_1, ..., w_{18}], w_i \in \mathcal{W}$. 

For the feed-forward network designed in IDGI, it uses a deep backbone network consisting of a series of residual blocks to encode the input image into high-level semantic feature maps ($1/64$ of the original input resolution) and then uses an extra fully connected (FC) layer to regress the latent vectors based on the last encoded features. However, a too deep network backbone may result in a mismatch between the feature used for inversion and the semantic level of StyleGAN.%, however, for the GAN inversion task there exists a best matching point of semantic level, which will be proved in ablation study.
Similarly, pSp also designs a deep backbone network but leverages the feature pyramid idea \cite{lin2017feature} to fuse the high-level feature with the low-level feature, then split each latent code into different feature pyramid levels. And for each latent code $w_i$ regression, an independent convolution based head (ConvHead) is used. Depending on the input feature resolution, the ConvHead consists of different numbers of convolutional layers that progressively downsample the features into $1\times1$ resolution, followed by one FC layer. Therefore, there are a total of 18 ConvHeads in pSp, thus introducing a lot of overheads. Besides, downsampling the features into $1\times 1$ will lose much useful information for predicting the latent code, thus resulting in unsatisfactory reconstruction quality. 

For more efficient network design, we ask three questions: ``\emph{Is there any guideline for the efficient backbone design?}'', ``\emph{How to split the latent code into different feature levels?}'', and ``\emph{Can we use cheaper regression heads?}''.

\vspace{0.5em}
\noindent\textbf{Shallower Backbone.} To answer the first question,  we propose a simple ``semantic level alignment'' principle, \ie, \emph{the maximum semantic level of the backbone network should match that of the target GAN model}. 
pSp \cite{Richardson2020EncodingIS} downsamples the resolution of the feature to $1/16$ of the input, and IDGI \cite{Zhu2020InDomainGI} downsamples the feature to $1/64$ of the input image. However, does the network really need to be such deep? In fact, there exists a semantic level best matching point. 
If the backbone network is too shallow, its encoded feature will not be able to predict the high-level latent codes well. On the other hand, too deep backbone network will not only introduce much overhead but also possibly bring even worse inversion results. 

To find the best match point, we follow the recognition network design convention \cite{szegedy2015going,he2016deep} and roughly regard the features of the same downsampled resolution as the same semantic level. Guided by this principle, we conduct a simple ablation on the backbone of pSp by using a single FC head like IDGI, and find the features corresponding to $1/8$ of the input resolution are sufficient to match the maximum semantic level of StyleGAN's latent codes and can even get better inversion results than the $1/16$ version. This demonstrates that \emph{ a deeper backbone network does not mean better inversion results}.
In contrast, the shallower encoder can not only bring the performance improvement but also reduce the model complexity. Moreover, we empirically find the feature pyramid does not help, and we guess it is because \wty{the features of deeper layers in the feature pyramid structure of pSp are derived from the features of early layers, and therefore the additional introduction of deeper-layer features in predicting low-dimensional latent codes does not bring extra information gain.}
%the low semantic-level latent codes are independent of high-level semantic information. 
Detailed results will be given in the ablation part. 

\vspace{0.5em}
\noindent\textbf{Hierarchical Structure.} Given the above backbone network, we get three different semantic levels of features (\ie, $1/8,1/4,1/2$). Inspired by pSp, we then split the features from different semantic levels to predict different parts of the latent codes. Considering the feature dimension of higher semantic levels are also higher, assigning more latent codes to higher semantic-level features will have a large model. To reduce the model size while maintaining a good inversion performance, we consider the above assignment problem as a constrained optimization problem.  Denote the latent code number assigned to each semantic level to be  $n_1, n_2, n_3$, it can be formally formulated as:
\begin{equation}
	\begin{aligned}
		n_1^*,n_2^*,n_3^*=&\underset{n_1,n_2,n_3}{\arg\max}\mathcal\quad {Q}(n_1,n_2,n_3) +\lambda \mathcal{P}(n_1,n_2,n_3), \\
		s.t., & \quad\quad n_{1}+n_{2}+n_{3} = 18,
	\end{aligned}
\end{equation} 
where $\mathcal{Q}, \mathcal{P}$ denotes the inversion quality and model size function with respect to $n_1,n_2,n_3$. Since it is hard to have an explicit function of $\mathcal{Q}$ due to its dependency on the training process, we simply adopt a simple binary search algorithm to find the rough optimal values. Specifically, we first regard $n_2,n_3$ as a whole and find the smallest $n_1$ with acceptable performance reduction (\wty{SSIM decreases by no more than $2\%$ of the original}), then we continue to find the smallest value for $n_2$ in a similar way. After the above search process, 
%$n_1, n_2, n_3$ equal to $9,5,4$ respectively.
we finally adopted a three-layer hierarchical structure to predict the front, middle and tail layers latent code, with the corresponding number of layers $n_1, n_2, n_3$ being $ 9 $, $ 5 $, and $ 4 $ respectively. %}

\vspace{0.5em}
\noindent\textbf{Efficient Prediction Head.} As mentioned above, there are 18 complex ConvHeads in pSp, each of which compresses the input feature into $1\times1$ resolution by a series of convolutional layers with stride of $ 2 $, and predict the single-layer latent code. We find such a complex ConvHead is not only  unnecessary but also introduces heavy overhead. To keep the regression head lightweight while keeping important information for regression, we design a very simple and efficient head, which just consists of an average pooling layer and a fully connected layer. For the deep ($1/8$), medium($1/4$), and shallow ($1/2$) features from the hierarchical structure, they will have such a simple head respectively. Besides, to balance performance and overhead, the average pooling layer in the three heads downsamples the input features to the resolution of $ 7\times7 $, $ 5\times5 $, and $ 3\times3 $ respectively, which effectively aggregates the information to predict the latent code while reducing the number of model parameters. 

\subsection{Loss Functions}
In order to train a better GAN inversion network, besides the commonly used losses, we improve the existing GAN inversion loss by introducing two new losses, \ie, multi-layer identity loss and multi-layer face parsing loss.

\vspace{0.5em}
\noindent\textbf{Common Losses.} The common losses consist of two types of constraints from the pixel level and the feature level, respectively. First, we use  $ \ell_2 $ loss to provide pixel-level supervision:
\begin{equation}
	\mathcal{L}_{2}=||\mathbf{x}-G(E(\mathbf{x}))||_{2},
\end{equation}
where $ \mathbf{x} $ represents the input image, $ E(\cdot) $ represents the GAN inversion network, $ G(\cdot) $ represents the StyleGAN, and $ G(E(\mathbf{x})) $ represents the reconstructed image of the input image. However, only using the $ \ell_2 $ loss will result in blurring reconstruction results, so we choose to use LPIPS \cite{Zhang2018TheUE} as our feature-level loss, which is demonstrated \cite{Guan2020CollaborativeLF} to yield clearer reconstruction results compared to the perceptual loss \cite{Johnson2016PerceptualLF}.
\begin{equation}
	\mathcal{L}_{LPIPS}=||F(\mathbf{x})-F(G(E(\mathbf{x})))||_{2},
\end{equation}
where $ F(\cdot) $ represents the AlexNet \cite{Krizhevsky2012ImageNetCW} feature extractor.

\vspace{0.5em}
\noindent\textbf{Multi-Layer Identity Loss.} As the key face image attribute, keeping the original identity information is extremely important for GAN inversion. Especially for human perception, whether the identity consistency between the inverted image and the original image can be effectively retained is a decisive factor for users to evaluate the quality of face inversion. Therefore, we leverage the multi-layer features from one pre-trained face recognition network (ArcFace \cite{Deng2019ArcFaceAA}) to impose semantic constraints on identity information. According to the resolution size, we choose $ 5 $ different levels of features as supervision to better supervise the semantic alignment of the identity information between the reconstructed image and the input image:
\begin{equation}
	\mathcal{L}_{m\_id}=\sum_{i=1}^5 (1-cos(R_{i}(\mathbf{x}), R_{i}(G(E(\mathbf{x}))))),
\end{equation}
where $cos$ means the cosine similarity and $ R_{i}(\mathbf{x}) $ denotes the feature corresponding to the $ i $-th semantic level from the face recognition network $ R $ of the input image $ \mathbf{x} $. 

\vspace{0.5em}
\noindent\textbf{Multi-Layer Face Parsing Loss.} Because features from the face recognition network often focus on capturing the global identity characteristics, relying on multi-layer identity loss alone cannot achieve accurate reconstructions of local details (\eg, glasses). To provide better local supervision, we introduce multiple layers of features from one pre-trained facial parsing network $ P $ \cite{FaceParsing} to provide a more localized knowledge:
\begin{equation}
	\mathcal{L}_{m\_par}=\sum_{i=1}^5 (1-cos(P_{i}(\mathbf{x}), P_{i}(G(E(\mathbf{x}))))),
\end{equation}

Similar to the well-known perceptual loss, even though the multi-layer identity loss and face parsing loss rely on extra pre-trained networks, we think it is valuable to introduce them into the GAN inversion to push the inversion quality to a new limit. More importantly, as they only appear in the training stage but not inference stage, they are indeed the free lunch for downstream applications. The generalization ability of the proposed loss to non-face domains will be verified in the experiment section.

\noindent To summarize, the overall loss function is defined as:
\begin{equation}
	\mathcal{L}=\lambda_{1}\mathcal{L}_{2}+\lambda_{2}\mathcal{L}_{LPIPS}+\lambda_{3}\mathcal{L}_{m\_id}+\lambda_{4}\mathcal{L}_{m\_par},
\end{equation}
where $ \lambda_{1} $, $ \lambda_{2} $, $ \lambda_{3} $, $ \lambda_{4} $ are set to $ 1 $, $ 0.8 $, $ 0.5 $, $ 1 $ respectively by default.
\subsection{Multi-Stage Refinement}

As shown in \cite{Zhu2020InDomainGI}, GAN inversion is a challenging problem and often difficult to achieve satisfactory inversion results with a single pass. Therefore, they first use a feed-forward network to get an initial inversion result, then refine it by using the optimization-based method as a post-processing step. Despite better inversion results, such optimization-based refinement is very time-consuming. But it motivates us to introduce  feed-forward-based recursive refinement. In this paper, thanks to our better inversion network design, we proposed the multi-stage refinement by purely using the feed-forward inversion networks, thus making the whole process still run in a feed-forward way. The overall framework diagram is shown in Figure \ref{fig:modelcompare} (c) and the whole process is formalized as follows:
\begin{equation}
	W_{t}=
	\begin{cases}
		E_{t}(\mathbf{x})& t=1\\
		E_{t}(\mathbf{x}, G(W_{t-1})) + W_{t-1}& t>1
	\end{cases}
\end{equation}
where $E_{t}$ represents the network of stage $t$ and $W_{t}=[w_1^t,...,w_{18}^t]$ is the inverted latent code. All the refinement stages $(t>1)$ adopt the same GAN inversion network structure as $E_1$ but take the concatenation of the original input image and the previous stage inverted image as input. Besides, the learning objective of the refinement stages is changed to predicting the residuals, so as to find a better latent code around the previous stage prediction result. 

To sum up, E2Style shares a similar spirit to IDGI's prediction-then-optimization approach. However, E2Style polishes the initial latent code by adding the multi-stage refinement, which requires only one forward pass to output the final result and does not bring a significant increase in time consumption while improving the inversion performance.

\section{Experiments}
In this section, we first compare E2Style quantitatively and qualitatively with the existing state-of-the-art approaches on the face domain. And the corresponding detailed ablation analysis is then provided to justify the effect of different improvement aspects. Finally, we demonstrate the generality of E2Style on non-face domains.

\subsection{Comparisons on the Face Domain}
\noindent\textbf{Implementation Details.} For the backbone structure, we follow the SE-ResNet50 \cite{Hu2020SqueezeandExcitationN} backbone but only keep the stages before the $1/16$ downsample stage. And we use StyleGAN2 pre-trained on the FFHQ dataset \cite{Karras2019ASG} as the target GAN model to inverse. In order to improve the generalization ability of the inversion network, the horizontal flip is used during training. Regarding the training strategy, we first train the first-stage network alone. After its convergence, we freeze it and then train the refinement stage network in a similar way. For all the network training, the base learning rate is set to $ 0.0001 $. By default, all network is trained for $ 25 $ epochs. Following pSp \cite{Richardson2020EncodingIS}, the Ranger optimizer is used, which is a combination of Rectified Adam \cite{Liu2020OnTV} with the Lookahead technique \cite{Zhang2019LookaheadOK}.

\vspace{0.5em}
\noindent\textbf{Datasets and Metrics.} To evaluate the cross-dataset inversion \wty{and editing} performance of existing methods, both the baselines and E2Style use the FFHQ dataset \cite{Karras2019ASG} with $ 70,000 $ faces as the training set, while the qualitative and quantitative comparisons are performed on the CelebA-HQ dataset \cite{Karras2018ProgressiveGO}. \wty{For inversion, }since the optimization-based method I2S \cite{Abdal2019Image2StyleGANHT} is too time-consuming, we randomly selected $ 2,000 $ images from the CelebA-HQ dataset and resized them to the resolution of $ 256\times256 $ for the evaluation of all methods. \wty{For editing, the entire CelebA-HQ dataset is used for quantitative and qualitative comparisons.} \wty{For the quantitative evaluation of the inversion}, five metrics are adopted: Peak Signal-to-Noise Ratio (PSNR) \cite{Hor2010ImageQM}, Structural SIMilarity (SSIM) \cite{Wang2004ImageQA}, IDentity Similarity (IDS), runtime (using a single NVIDIA GEFORCE RTX 3090 GPU), and model size. \wty{For quantitative evaluation of editing quality, we use FID \cite{Heusel2017GANsTB} and IDentity Similarity (IDS) as metrics and conduct a user study.} For SSIM, PSNR, and IDS, higher indicates better. The method we used here to calculate the identity similarity is Curricularface \cite{Huang2020CurricularFaceAC}, instead of ArcFace \cite{Deng2019ArcFaceAA} which we used for training.

\begin{figure*}[htp]
	\begin{center}
		\setlength{\tabcolsep}{1pt}
		\begin{tabular}{cccccccc}
			\footnotesize{{pSp \cite{Richardson2020EncodingIS}}}& \footnotesize{{e4e \cite{Tov2021DesigningAE}}}  & \footnotesize{{ReStyle \cite{alaluf2021restyle}}} & \footnotesize{{DNI \cite{Yang2021ADN}}} & \footnotesize{{I2S \cite{Abdal2019Image2StyleGANHT}}} & \footnotesize{{E2Style-1Stage}} & \footnotesize{{E2Style-2Stage}} & \footnotesize{{Input Image}}
			\\
			\includegraphics[width=2.18cm]{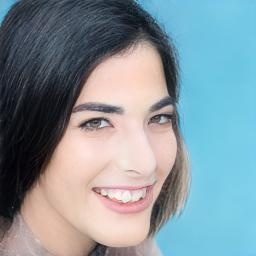}
			&\includegraphics[width=2.18cm]{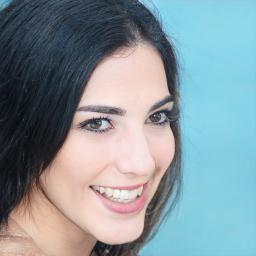}
% 			&\includegraphics[width=2.18cm]{images/inversioncompare-image/IDGI_1.png}
            &\includegraphics[width=2.18cm]{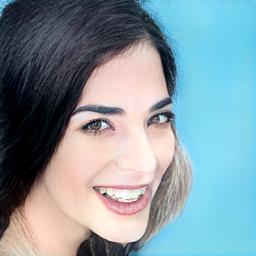}
			&\includegraphics[width=2.18cm]{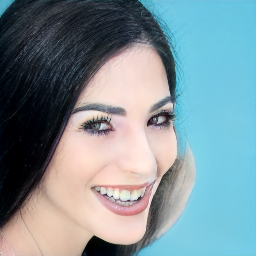}
			&\includegraphics[width=2.18cm]{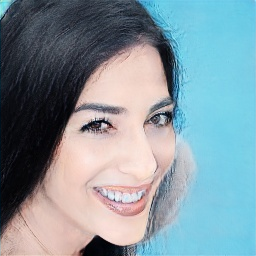}
			&\includegraphics[width=2.18cm]{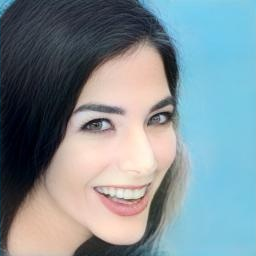}
			&\includegraphics[width=2.18cm]{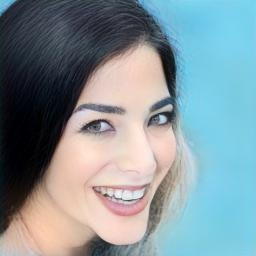}
			&\includegraphics[width=2.18cm]{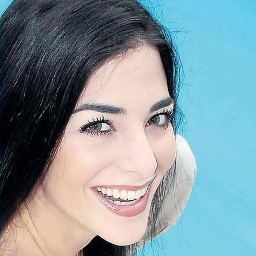}
			\\
			
			\includegraphics[width=2.18cm]{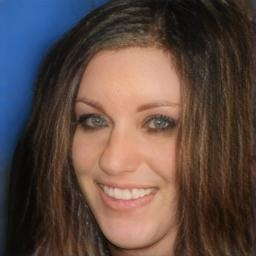}
			&\includegraphics[width=2.18cm]{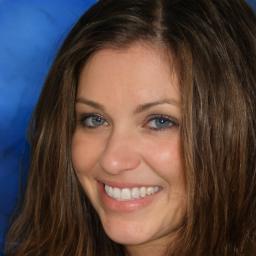}
% 			&\includegraphics[width=2.18cm]{images/inversioncompare-image/IDGI_2.png}
            &\includegraphics[width=2.18cm]{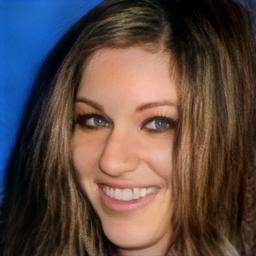}
			&\includegraphics[width=2.18cm]{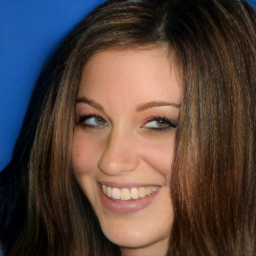}
			&\includegraphics[width=2.18cm]{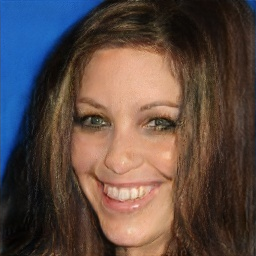}
			&\includegraphics[width=2.18cm]{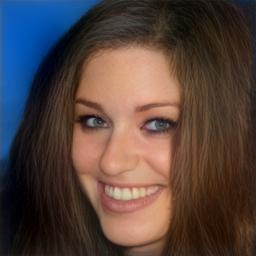}
			&\includegraphics[width=2.18cm]{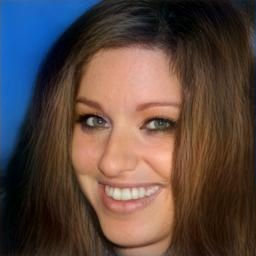}
			&\includegraphics[width=2.18cm]{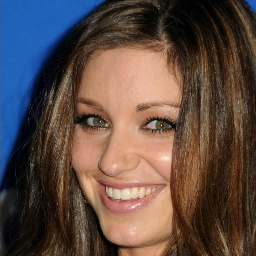}
			\\
			
			\includegraphics[width=2.18cm]{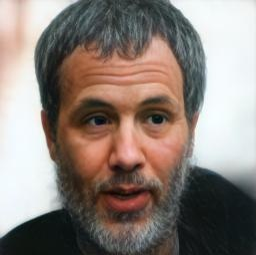}
			&\includegraphics[width=2.18cm]{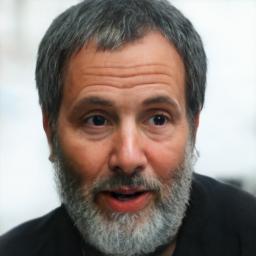}
% 			&\includegraphics[width=2.18cm]{images/inversioncompare-image/IDGI_3.png}
            &\includegraphics[width=2.18cm]{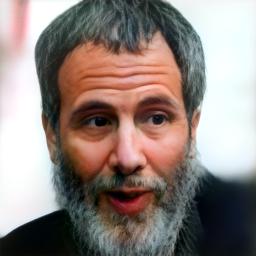}
			&\includegraphics[width=2.18cm]{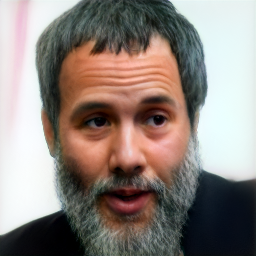}
			&\includegraphics[width=2.18cm]{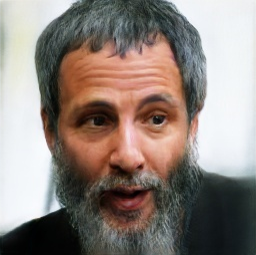}
			&\includegraphics[width=2.18cm]{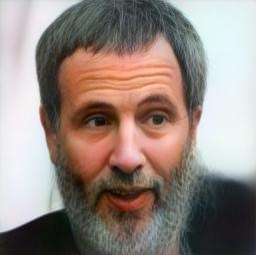}
			&\includegraphics[width=2.18cm]{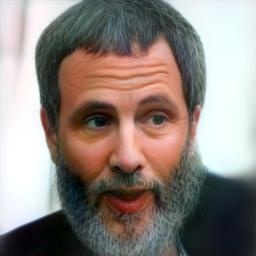}
			&\includegraphics[width=2.18cm]{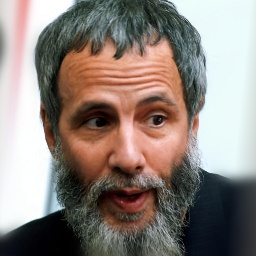}
			\\  
			
		\end{tabular}
	\end{center}
	\caption{Visual comparison of the GAN inversion quality on the CelebA-HQ dataset. \wty{Our E2Style performs better than existing forward-based methods and comparably to state-of-the-art optimization-based methods while maintaining high efficiency as well as forward-based methods.}} 
	\label{fig:inversioncompare}
\end{figure*}

\begin{table}[t]
	\caption{Quantitative comparison \wty{of inversion quality} on the CelebA-HQ dataset, IDS means identity similarity, $*$ means without residual learning, \aq{C means replacing the average pooling layer with continuous convolutional layers}. Our E2Style achieves comparable results to optimization-based methods and better results than feed-forward-based methods while maintaining high efficiency.}	
	\label{tab:quantitativecompare}
	\centering
	\small
	\setlength{\tabcolsep}{0.4em}{
		\begin{tabular}{lccccc}
			\hline
			Methods & SSIM & PSNR & IDS & RunTime(s) & Param(M)\\
			\hline
			pSp  & 0.58 & 20.7 & 0.57 & 0.09 & 262\\
			IDGI-Encoder  & 0.51 & 18.9 & 0.19 & \textbf{0.02} & 165\\
			ReStyle & 0.62 & 21.6 & 0.67 & 0.36 & 201\\
			\wty{e4e} & 0.55 & 19.5 & 0.51 & 0.09 & 262\\
			E2Style-1Stage & 0.63 & 21.3 & 0.69 & 0.04 & \textbf{85}\\
			\aq{E2Style-1Stage-C} & 0.62 & 21.2 & 0.68 & 0.04 & 87\\
			E2Style-2Stage & 0.66 & 22.5 & 0.74 & 0.09 & 170\\
			E2Style-3Stage & \textbf{0.67} & \textbf{23.0} & \textbf{0.75} & 0.13  & 255\\
			E2Style-2Stage$*$ & 0.63 & 21.3 & 0.69 & 0.09 & 170\\
			\hline
			IDGI & 0.62 & 22.3 & 0.37 & 6.60 & 165\\
			DNI & 0.65 & 21.0 & 0.55 & 13.1 & 525 \\
			I2S & 0.67 & 23.9 & 0.60 & 54.4 & -\\
			
			\hline
		\end{tabular}
	}

\end{table}

\noindent\textbf{\wty{Inversion Quality}.} We compare E2Style with the current state-of-the-art GAN inversion approaches, including forward-based methods: pSp \cite{Richardson2020EncodingIS}, IDGI-Encoder \cite{Zhu2020InDomainGI}, ReStyle \cite{alaluf2021restyle}, \wty{e4e \cite{Tov2021DesigningAE}}, optimization-based method: I2S \cite{Abdal2019Image2StyleGANHT}, and hybrid method: IDGI \cite{Zhu2020InDomainGI}, DNI \cite{Yang2021ADN}.
For ReStyle, the iterative refinement mechanism is executed 5 times as they recommend.
For IDGI, we followed the procedure in their paper: the output of the network was used as the initialization point and the optimization is iterated $ 100 $ times. I2S used the mean latent code as initialization and then iterated $ 1,000 $ times for each input image. For DNI, the latent code is initialized by the domain-guided encoder, and then the noise optimization process is iterated 100 times. %Note that since DNI requires 16G of GPU memory to run, we measured its runtime using the single NVIDIA GeForce RTX 3090, which is already marked by italics in Table \ref{tab:quantitativecompare}.

As shown in Table \ref{tab:quantitativecompare}, our single-stage inversion network already outperforms all existing single-stage forward-based methods with the smaller model size. With multi-stage refinement, our 2-stage network outperforms the multi-stage forward-based method ReStyle, the hybrid method IDGI, DNI. Our 3-stage network has comparable performance with the optimization-based method I2S in terms of SSIM metric. Moreover, our identity consistency and speed both have obvious advantages over I2S. We further provide some qualitative comparison examples in Figure \ref{fig:inversioncompare}, whose quality rank is consistent with the quantitative results.

\noindent\textbf{\wty{Editing Quality.}} \wty{We now compare the editing quality of E2Style with baselines. InterFaceGAN \cite{Shen2020InterpretingTL} is used to obtain editing directions, and then these are used to edit latent codes obtained by different inversion methods. As shown in Figure \ref{fig:semanticfaceediting}, compared to pSp \cite{Richardson2020EncodingIS}, e4e \cite{Tov2021DesigningAE}, Restyle \cite{alaluf2021restyle}, our results are more desirable. For example, when editing the expression, our method better preserves the identity consistency of the image. In terms of age manipulation, our method manipulates other facial semantics less while changing the age of the target face.} 

\wty{In Table \ref{tab:editing-quantitativecompare}, we provide quantitative comparisons. FID and IDentity Similarity are calculated between input images and edited images. E2Style achieves the best identity consistency while completing the editing. To perceptually measure the performance of different methods for editing, we ask \aq{200} volunteers to conduct a user study with \aq{20} randomly picked result groups, and ask them to select the one which completes the specified editing while retaining the other irrelevant attributes to a large extent. The preference rates are shown in Table \ref{tab:editing-quantitativecompare} and demonstrate the superiority of our method.}

\begin{table}[t]
	\caption{\wty{Quantitative comparison of editing quality on the CelebA-HQ dataset, IDS means identity similarity. Our E2Style achieves the best identity consistency and highest preference rates.}}	
	\label{tab:editing-quantitativecompare}
	\centering
	\small
	\setlength{\tabcolsep}{1.56em}{
		\begin{tabular}{lccc}
			\hline
			Methods & FID & IDS & Preference Rate \\
			\hline
			pSp  & 54.4 & 0.33 & 12.0\% \\
			ReStyle & \textbf{48.8} & 0.40 & 14.8\% \\
			e4e & 55.5 & 0.31 & 18.8\% \\
			E2Style   & 49.4 & \textbf{0.49} & \textbf{54.5\%} \\
			\hline
		\end{tabular}
	}
	
\end{table}

\begin{figure*}[t]
	\begin{center}
		\setlength{\tabcolsep}{0.5pt}
		\begin{tabular}{m{0.3cm}<{\centering}m{1.7cm}<{\centering}m{1.7cm}<{\centering}m{1.7cm}<{\centering}m{1.7cm}<{\centering}m{1.7cm}<{\centering}m{1.7cm}<{\centering}m{1.7cm}<{\centering}m{1.7cm}<{\centering}m{1.7cm}<{\centering}m{1.7cm}<{\centering}}
			& \small{Original} & \multicolumn{2}{c}{\small{Expression}} & \multicolumn{2}{c}{\small{Age}} & \small{Original} & \multicolumn{2}{c}{\small{Expression}} & \multicolumn{2}{c}{\small{Age}}
			\\
			\raisebox{0.6cm}{\rotatebox[origin=c]{90}{\footnotesize{{Restyle \cite{alaluf2021restyle}}}}}
			&\includegraphics[width=1.68cm]{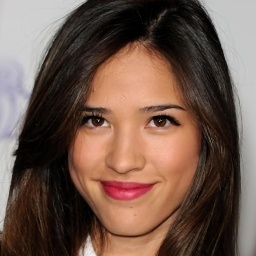}
			&\includegraphics[width=1.68cm]{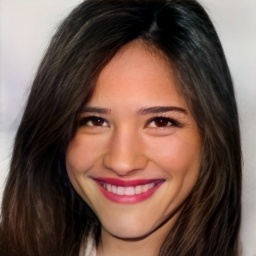}
			&\includegraphics[width=1.68cm]{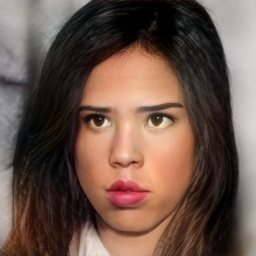}
			&\includegraphics[width=1.68cm]{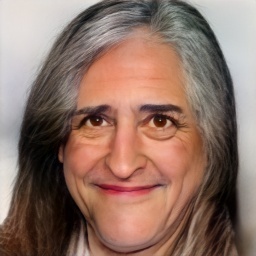}
			&\includegraphics[width=1.68cm]{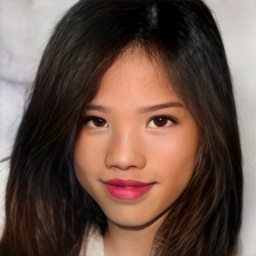}
			&\includegraphics[width=1.68cm]{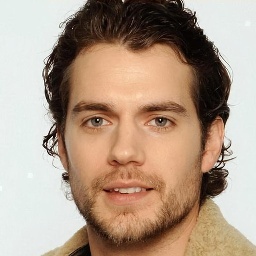}
			&\includegraphics[width=1.68cm]{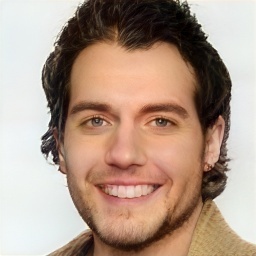}
			&\includegraphics[width=1.68cm]{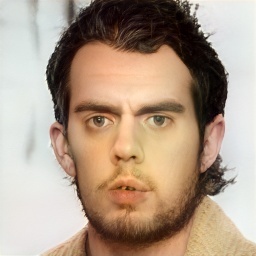}
			&\includegraphics[width=1.68cm]{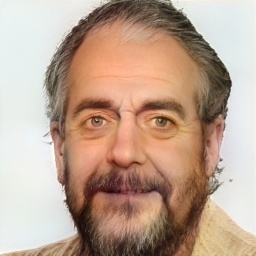}
			&\includegraphics[width=1.68cm]{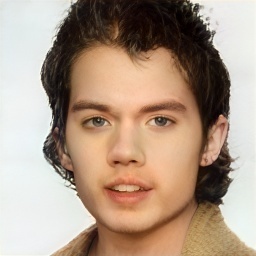}
			\\
			
			\raisebox{0.39cm}{\rotatebox[origin=c]{90}{\footnotesize{{e4e \cite{Tov2021DesigningAE}}}}}
			&\includegraphics[width=1.68cm]{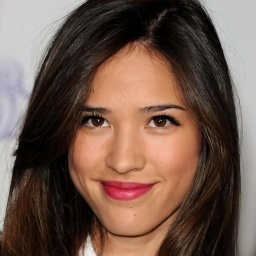}
			&\includegraphics[width=1.68cm]{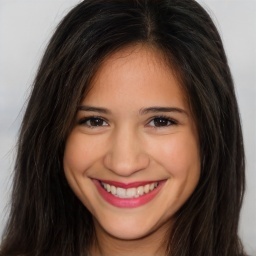}
			&\includegraphics[width=1.68cm]{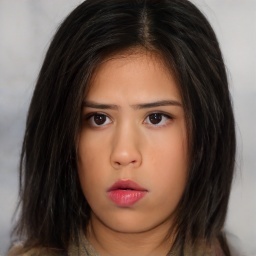}
			&\includegraphics[width=1.68cm]{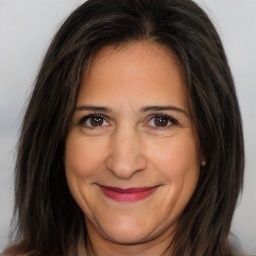}
			&\includegraphics[width=1.68cm]{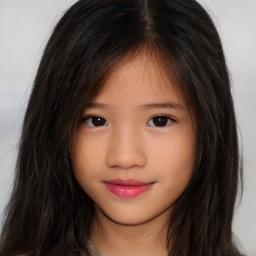}
			&\includegraphics[width=1.68cm]{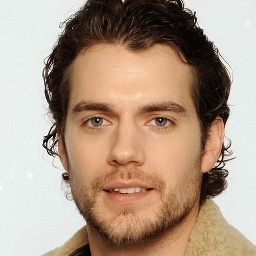}
			&\includegraphics[width=1.68cm]{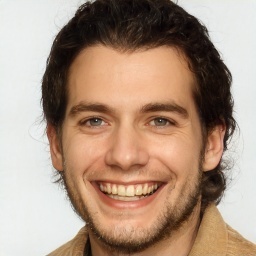}
			&\includegraphics[width=1.68cm]{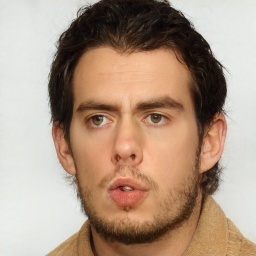}
			&\includegraphics[width=1.68cm]{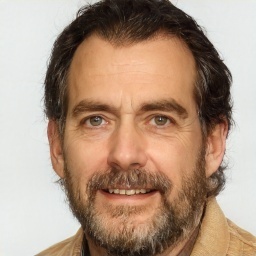}
			&\includegraphics[width=1.68cm]{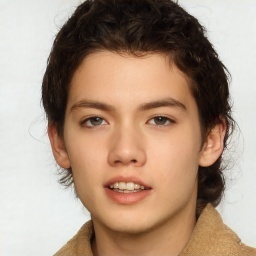}
			\\
			%			\small{pSp \cite{Richardson2020EncodingIS}}
			\raisebox{0.5cm}{\rotatebox[origin=c]{90}{\footnotesize{{pSp \cite{Richardson2020EncodingIS}}}}}
			&\includegraphics[width=1.68cm]{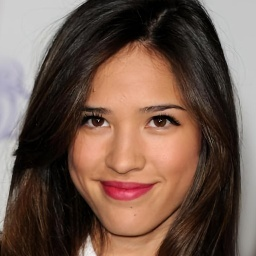}
			&\includegraphics[width=1.68cm]{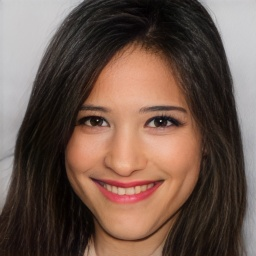}
			&\includegraphics[width=1.68cm]{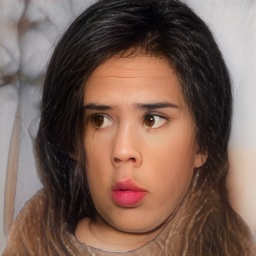}
			&\includegraphics[width=1.68cm]{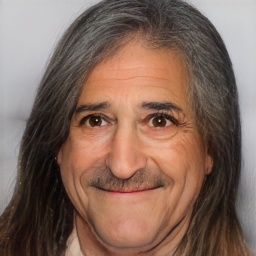}
			&\includegraphics[width=1.68cm]{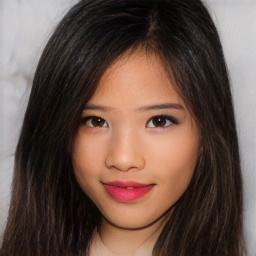}
			&\includegraphics[width=1.68cm]{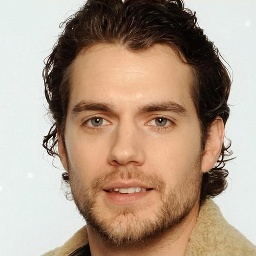}
			&\includegraphics[width=1.68cm]{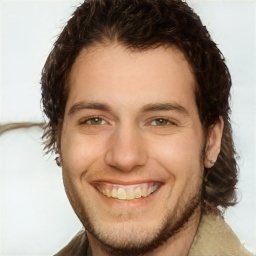}
			&\includegraphics[width=1.68cm]{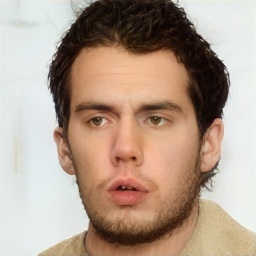}
			&\includegraphics[width=1.68cm]{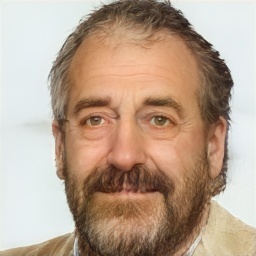}
			&\includegraphics[width=1.68cm]{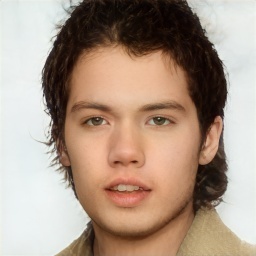}
			\\
			%			\small{Ours}
			\raisebox{0.5cm}{\rotatebox[origin=c]{90}{\footnotesize{{E2Style}}}}
			&\includegraphics[width=1.68cm]{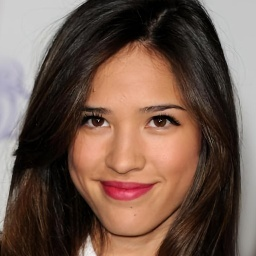}
			&\includegraphics[width=1.68cm]{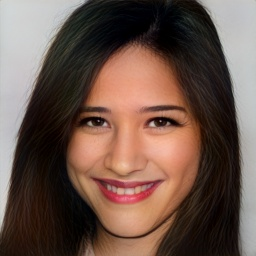}
			&\includegraphics[width=1.68cm]{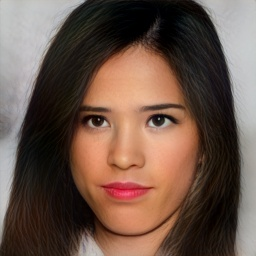}
			&\includegraphics[width=1.68cm]{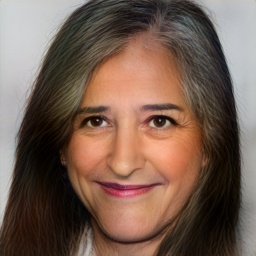}
			&\includegraphics[width=1.68cm]{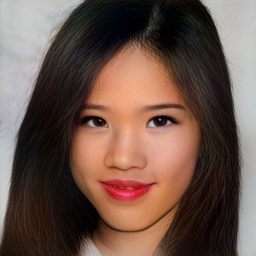}
			&\includegraphics[width=1.68cm]{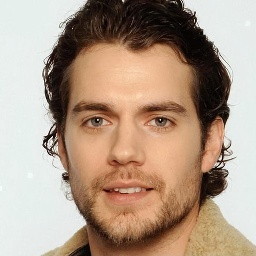}
			&\includegraphics[width=1.68cm]{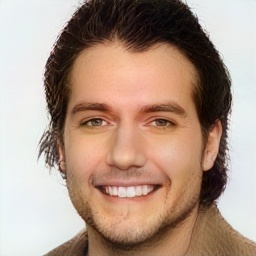}
			&\includegraphics[width=1.68cm]{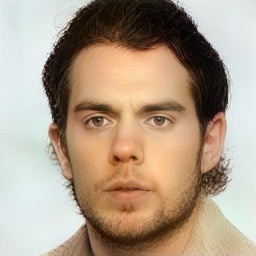}
			&\includegraphics[width=1.68cm]{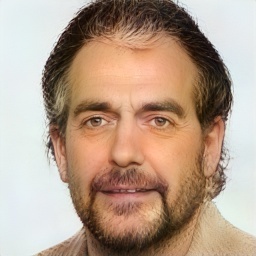}
			&\includegraphics[width=1.68cm]{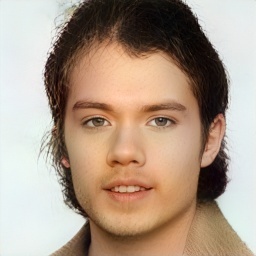}
			\\
			
		\end{tabular}
	\end{center}
	
	\caption{\wty{Qualitative comparison of our E2Style with pSp \cite{Richardson2020EncodingIS}, e4e \cite{Tov2021DesigningAE}, and Restyle \cite{alaluf2021restyle} on editing. As can be seen, our approach shows the best identity consistency while completing the editing.}} 
	\label{fig:semanticfaceediting}
	
\end{figure*}

\subsection{Ablation Study}

\begin{table}[t]
	\caption{Quantitative ablation study. Size represents the maximum factor to downsample the input image. For the prediction head, 1E, 3E, and 18C represent the single Efficient Head, the three-level Efficient Heads, and 18 independent ConvHeads, respectively. Regarding training loss, C, SI, MI, and MP stand for common, single-layer identity, multi-layer identity, and multi-layer parsing loss, respectively.}	
	\label{tab:ablation}
	\centering
	\small
	\setlength{\tabcolsep}{3.5pt}
	\begin{tabular}{l|lcc|cccc}
		\hline
		& Size & Head & Loss & SSIM & PSNR & IDS &Param(M)\\
		\hline
		(a) & $ \downarrow4 $ & 1E & C+SI  & 0.57 & 20.3 & 0.62 & 59\\
		(b) & $ \downarrow16 $ & 1E & C+SI  & 0.57 & 20.5 & 0.66 & 262\\
		(c) & $ \downarrow8 $ & 1E & C+SI  &0.59 & 20.9 & 0.67 & 133\\
		\textbf{(d)} & $ \downarrow8 $ & 3E & C+SI  & 0.59 & 21.0 & 0.67 & 85\\
		(e) & $ \downarrow8 $ & 18C & C+SI  & 0.55 & 20.0 & 0.61 & 233\\
		\hline
		(f) & $ \downarrow8 $ & 3E & C  & 0.59 & 21.0 & 0.27 & 85\\
		(g) & $ \downarrow8 $ & 3E & C+MI  & 0.60 & 20.9 & 0.69 & 85\\
		\textbf{(h)} & $ \downarrow8 $ & 3E & C+MI+MP  & 0.63 & 21.3 & 0.69 & 85\\
		\hline
	\end{tabular}

\end{table}
\vspace{0.5em}
\noindent\textbf{Effectiveness of Network Structure Design.} In the top part of Table \ref{tab:ablation}, we first validate the effectiveness of the proposed network structure design. (a), (b), and (c) indicate that we used the single efficient prediction head to directly predict the latent code after downsampling the input image to $ 1/4$, $ 1/16 $, and $ 1/8 $ of the original resolution, respectively. 
For all these three settings we used common ($ \ell_2 $, LPIPS) and the single-layer identity loss function to train the network.
It clearly shows that deeper networks are not better, and the larger network capacity does not lead to a performance gain. For the inversion task, there is a point where the semantic level of StyleGAN's latent code is best matched, which may explain the unsatisfactory performance of the IDGI encoder. 

It is intuitive to design the inversion network to have the same stage number as StyleGAN and strictly match their semantic levels based on the feature resolution, but our experimental results above show deeper inversion networks even have worse performance. It also indicates that fully relying on the feature resolution for semantic alignment is not a good choice. Moreover, that deep inversion network will be super large and infeasible for real-time applications. Considering these points, we design a relatively shallow inversion network and 
%In order to achieve a more efficient model without degrading the performance, we 
adopt a three-layer hierarchical structure to predict the latent code as (d). For (e), we took the same scheme as pSp \cite{Richardson2020EncodingIS}, \ie, we used 18 Convheads to predict the latent code of each layer. Compared with (d), (e) brings a larger number of parameters but worse performance. We guess this may be because continuous downsampling to the feature resolution of $ 1\times1 $ will lead to excessive information missing, while using average pooling can effectively aggregate useful information with a shorter path.

\begin{figure}[t]
	\begin{center}
		\setlength{\tabcolsep}{1pt}
		\begin{tabular}{cccc}
			\footnotesize{{Input Image}} & \footnotesize{{w/o ID}} & \footnotesize{{w/ Single-ID}}  & \footnotesize{{w/ Multi-ID}}
			\\
			\includegraphics[width=2.1cm]{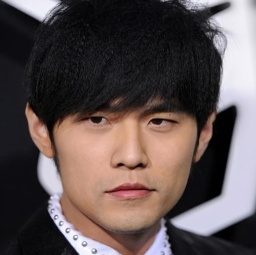}
			&\includegraphics[width=2.1cm]{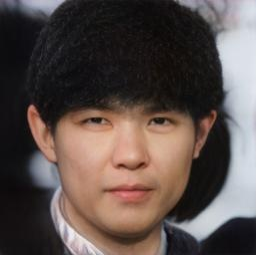}
			&\includegraphics[width=2.1cm]{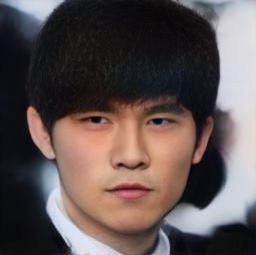}
			&\includegraphics[width=2.1cm]{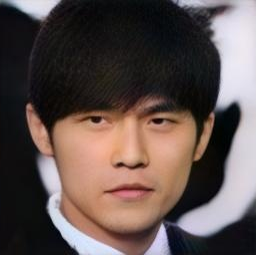}
			\\
			
			\includegraphics[width=2.1cm]{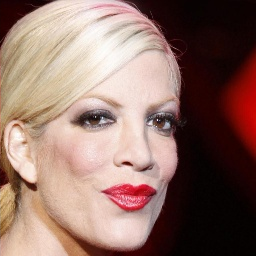}
			&\includegraphics[width=2.1cm]{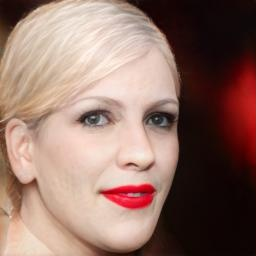}
			&\includegraphics[width=2.1cm]{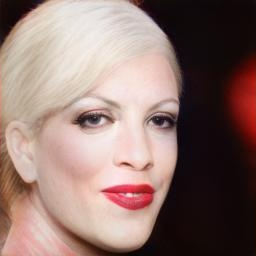}
			&\includegraphics[width=2.1cm]{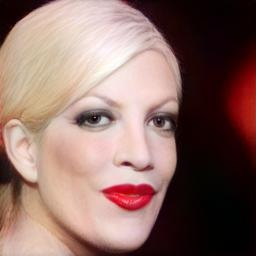}
			
			\\
		\end{tabular}
	\end{center}
	\caption{The effect of multi-layer identity loss. With the multi-layer identity loss, the identity consistency between the reconstructed image and the input image is significantly improved.} 
	\label{fig:ablation-id}
\end{figure}

\begin{figure}[h]
	\begin{center}
		\setlength{\tabcolsep}{5pt}
		\begin{tabular}{ccc}
			\footnotesize{{Input Image}} & \footnotesize{{w/o Multi-Parsing}} & \footnotesize{{w/ Multi-Parsing}}
			\\
			\includegraphics[width=2.60cm]{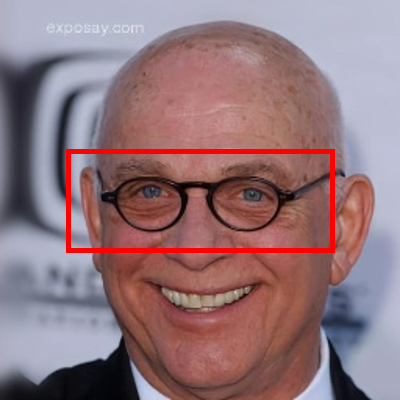}
			&\includegraphics[width=2.60cm]{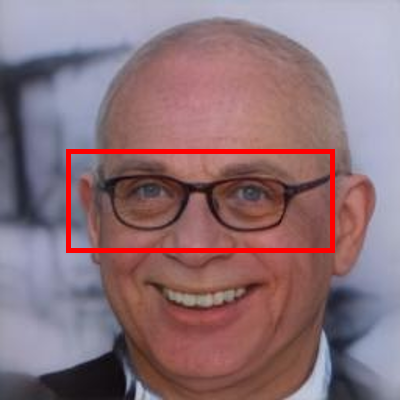}
			&\includegraphics[width=2.60cm]{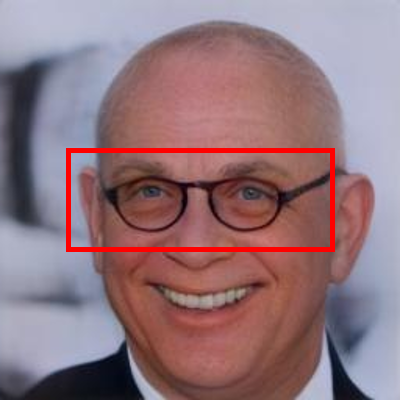}
			\\
			\includegraphics[width=2.60cm]{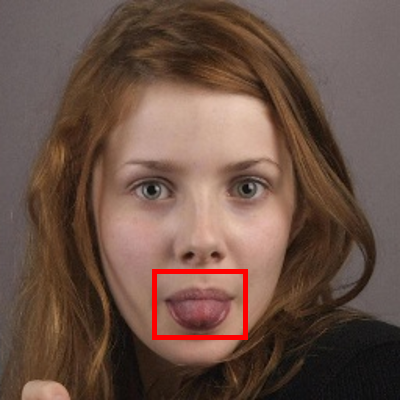}
			&\includegraphics[width=2.60cm]{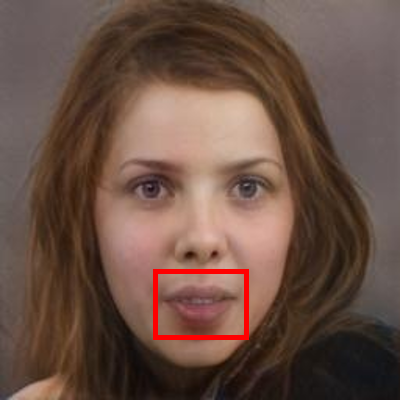}
			&\includegraphics[width=2.60cm]{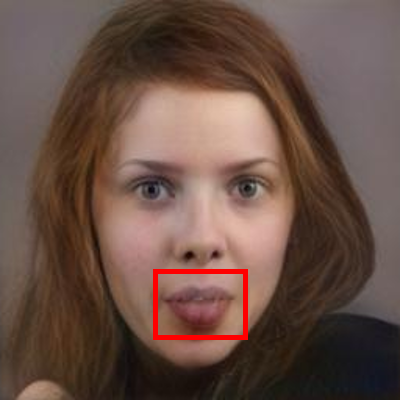}
			\\
			
		\end{tabular}
	\end{center}
	\caption{The effect of multi-layer face parsing loss. With the multi-layer face parsing loss, the network is able to pay attention to the local face regions, thus more accurately reconstructing the local details, \eg, the glasses and tongue of the two cases.} 
	\label{fig:ablation-parse}
\end{figure}

Next, we give the quantitative ablation study about feature pyramid network (FPN). Compared with baseline (d), we add the same feature pyramid structure as pSp \cite{Richardson2020EncodingIS} to (d), however, quantitative results (exactly the same SSIM, PSNR, and IDS) prove that this structure does not lead to performance gains. But model parameters are increased from 85M to 98M and the running time is increased from 0.04 seconds to 0.05 seconds. \wty{We guess it is because the features of deeper layers in the feature pyramid structure of pSp are derived from the features of early layers, and therefore the additional introduction of deeper-layer features in predicting low-dimensional latent codes does not bring extra information gain.}
%We guess it is because low semantic-level latent codes are independent of high-level semantic information.

% we demonstrate that the average pooling layer we used achieves a better balance between performance and overhead with respect to convolutional layers. Here 

\aq{Finally, we provide an experiment E2Style-1Stage-C which replaces the average pooling layer in E2Style-1Stage with continuous convolutional layers to downsample the features to the same size as E2Style-1Stage. The quantitative results in Table \ref{tab:quantitativecompare} show that more parameters do not result in performance improvement. Therefore, we choose the more efficient and parameter-free average pooling to downsample the features.}

\begin{table*}[t]
	\caption{Quantitative comparison of different encoder structures on the non-face domains. Note that pSp, IDGI and E2Style-1Stage all use the same training losses and strategies, the only difference between them is the network structure. It is clear that our network structure achieves the best performance with the lowest number of network parameters.}	
	\label{tab:otherdataset}
	\centering
	\small
	\setlength{\tabcolsep}{0.55em}{
		\begin{tabular}{l | cccc | cccc | cccc}
			\hline
			\multicolumn{1}{ c }{{}}	& \multicolumn{4}{ c }{{Cats}} & \multicolumn{4}{ c }{{Horses}} & \multicolumn{4}{ c }{{Cars}} \\
			\hline
			Methods & SSIM & PSNR & Time(s) & Param(M) & SSIM & PSNR & Time(s) & Param(M) & SSIM & PSNR & Time(s) & Param(M)\\
			\hline
			pSp  & 0.39 & 18.1 & 0.06 & 206 & 0.33 & 17.1 & 0.06 & 206 & 0.40 & 16.2 & 0.06 & 234\\
			IDGI  & 0.40 & 18.4 & \textbf{0.02} & 165 & 0.34 & 17.7 & \textbf{0.02} & 165 & 0.41 & 16.7 & \textbf{0.03} & 201\\
			E2Style-1Stage & \textbf{0.41} & \textbf{18.5} & \textbf{0.02} & \textbf{85} & \textbf{0.37} & \textbf{17.8} & \textbf{0.02} & \textbf{85} & \textbf{0.43} & \textbf{16.8} & \textbf{0.03} & \textbf{95}\\
			\hline
		\end{tabular}
	}
	
\end{table*}

\begin{figure*}[t]
	\begin{center}
		\setlength{\tabcolsep}{1pt}
		\begin{tabular}{cccccccc}
			\footnotesize{{pSp \cite{Richardson2020EncodingIS}}} &\footnotesize{{IDGI \cite{Zhu2020InDomainGI} }} & \footnotesize{{E2Style-1Stage}} & \footnotesize{{Input Image}} & \footnotesize{{pSp \cite{Richardson2020EncodingIS}}} &\footnotesize{{IDGI \cite{Zhu2020InDomainGI}}}  & \footnotesize{{E2Style-1Stage}} & \footnotesize{{Input Image}}
			\\
			\includegraphics[width=2.18cm]{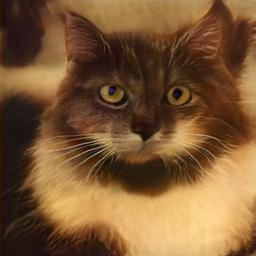}
			&\includegraphics[width=2.18cm]{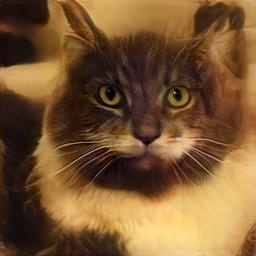}
			&\includegraphics[width=2.18cm]{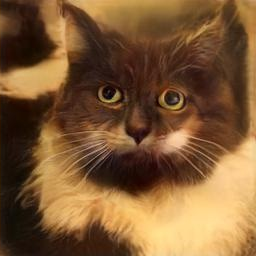}
			&\includegraphics[width=2.18cm]{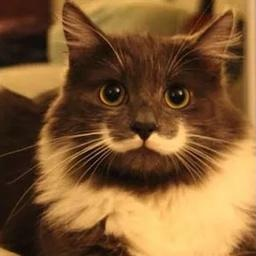}
			&\includegraphics[width=2.18cm]{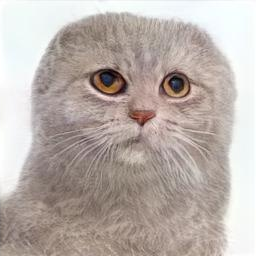}
			&\includegraphics[width=2.18cm]{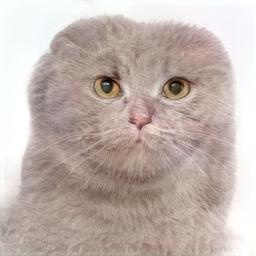}
			&\includegraphics[width=2.18cm]{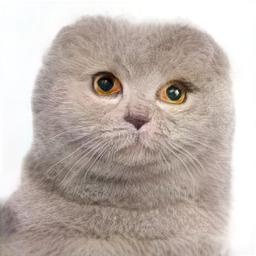}
			&\includegraphics[width=2.18cm]{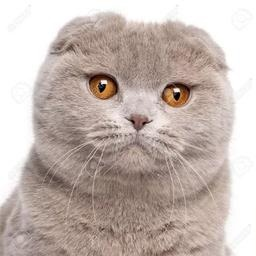}
			\\
			
			\includegraphics[width=2.18cm]{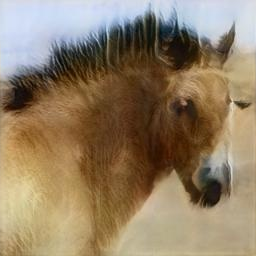}
			&\includegraphics[width=2.18cm]{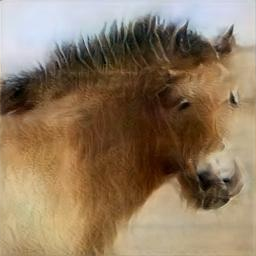}
			&\includegraphics[width=2.18cm]{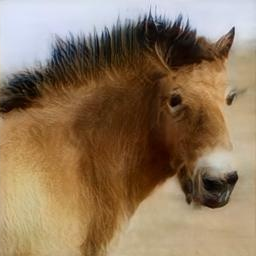}
			&\includegraphics[width=2.18cm]{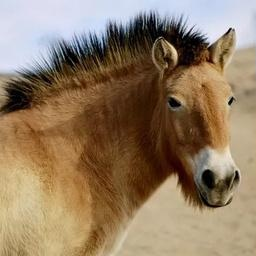}
			&\includegraphics[width=2.18cm]{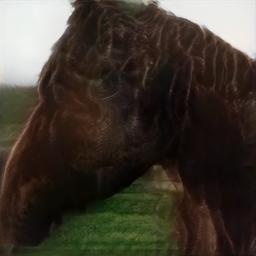}
			&\includegraphics[width=2.18cm]{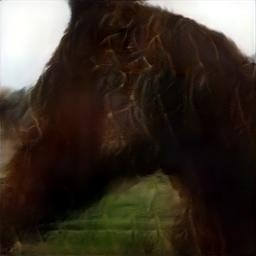}
			&\includegraphics[width=2.18cm]{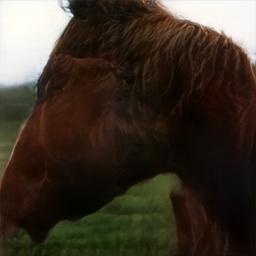}
			&\includegraphics[width=2.18cm]{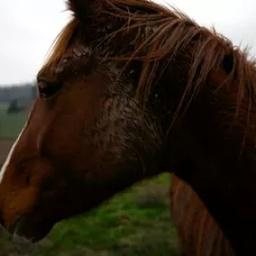}
			\\
			
			\includegraphics[width=2.18cm]{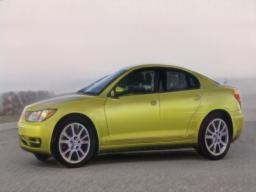}
			&\includegraphics[width=2.18cm]{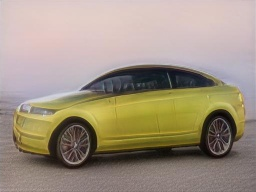}
			&\includegraphics[width=2.18cm]{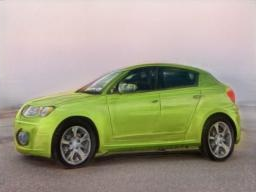}
			&\includegraphics[width=2.18cm]{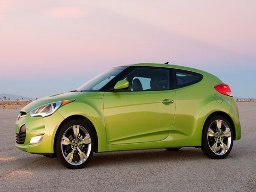}
			&\includegraphics[width=2.18cm]{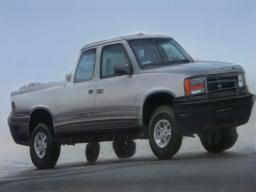}
			&\includegraphics[width=2.18cm]{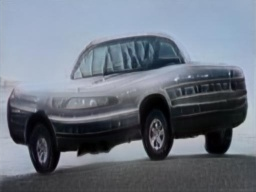}
			&\includegraphics[width=2.18cm]{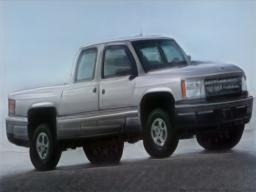}
			&\includegraphics[width=2.18cm]{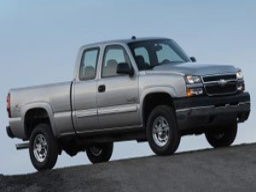}
			\\
			
		\end{tabular}
	\end{center}
	\caption{Visual comparison of the GAN inversion quality on the non-face domains. Thanks to our improvements in network structure, our encoder generates higher quality reconstruction results with the lowest network parameters compared to pSp, IDGI.} 
	\label{fig:otherdatasetfigure}
\end{figure*}

\vspace{0.5em}

\noindent\textbf{Importance of Multi-Layer Identity and Parsing Losses.} The bottom part of Table \ref{tab:ablation} shows our quantitative results for removing identity loss, introducing multi-layer identity loss, and introducing both multi-layer identity loss and multi-layer face parsing loss on top of (d), respectively. It well demonstrates the effectiveness of multi-layer identity and face parsing loss. We further provide some qualitative visual comparisons in Figure \ref{fig:ablation-id} and Figure \ref{fig:ablation-parse}. With the multi-layer identity loss, the identity consistency between the reconstructed image and the input image is significantly improved. With the multi-layer face parsing loss, the network is able to pay attention to the local face regions, thus more accurately reconstructing the local details, \eg, the glasses and tongue of the two cases.

\vspace{0.5em}
\noindent\textbf{Significance of the Refinement Stage.} As shown in Table \ref{tab:quantitativecompare}, our 2-stage inversion network significantly improves all the metrics compared to the 1-stage counterpart, but the gain starts to saturate by adding more stages. Considering both speed and performance. we adopt the two-stage inversion network as the default setting. In the last row of Table \ref{tab:quantitativecompare}, we further change the second stage learning objective to directly predict the latent code instead of the residuals of the first stage latent code. The experimental results show that learning the residuals in the refinement stages is extremely important. In contrast, if we continue to learn the absolution latent code, the refinement stage is still difficult to bring better performance, which may be because the latent code residuals (smaller value range) have smaller variance than the absolute latent codes, thus making the inversion network easier to regress.

\begin{figure}[t]
	\begin{center}
		\setlength{\tabcolsep}{5pt}
		\begin{tabular}{ccc}
			\footnotesize{{Input Image}} & \footnotesize{{w/o Multi-Seg}} & \footnotesize{{w/ Multi-Seg}}
			\\
			\includegraphics[width=2.60cm]{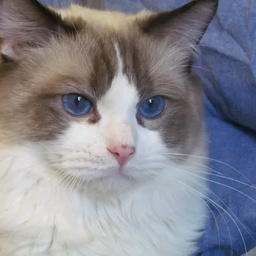}
			&\includegraphics[width=2.60cm]{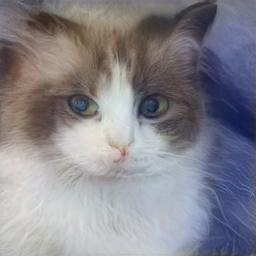}
			&\includegraphics[width=2.60cm]{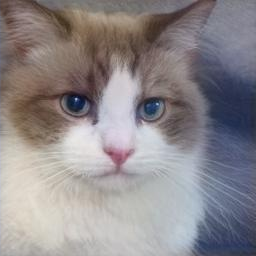}
			\\
			\includegraphics[width=2.60cm]{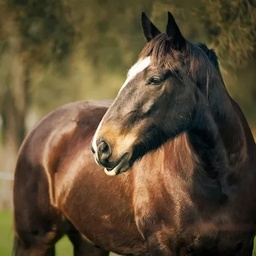}
			&\includegraphics[width=2.60cm]{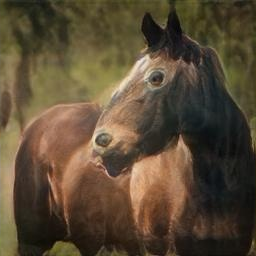}
			&\includegraphics[width=2.60cm]{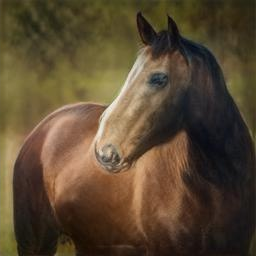}
			\\
			\includegraphics[width=2.60cm]{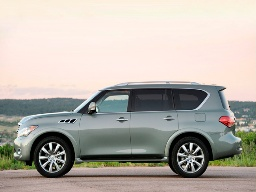}
			&\includegraphics[width=2.60cm]{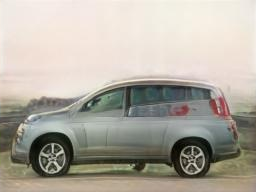}
			&\includegraphics[width=2.60cm]{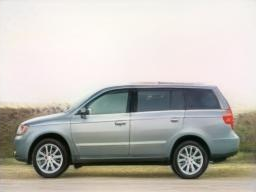}
			\\			
			
		\end{tabular}
	\end{center}
	\caption{The effect of multi-layer segmentation loss. It enables the network to pay more attention to salient objects that are of more interest to humans, resulting in a large improvement in perceptual quality.} 
	\label{fig:seg-ablation}
\end{figure}

\subsection{Comparisons on Non-Face Domains}
We will verify the versatility of the proposed improvements on the network design and loss function to non-face domains.
\vspace{0.5em}

\begin{figure*}[t]
	\centering
	\includegraphics[width=\textwidth]{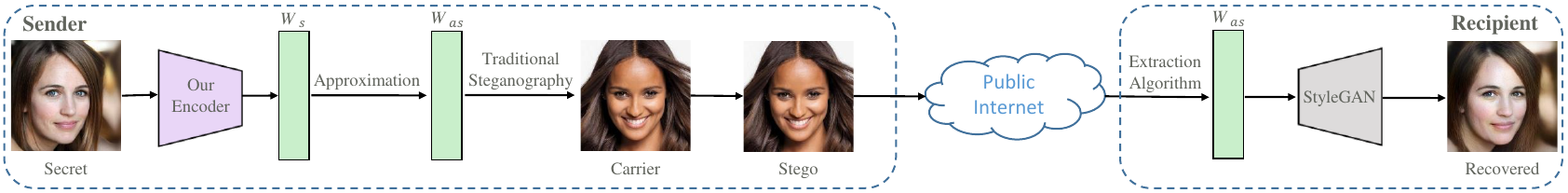} 
	\caption{The illustration of our proposed secure deep hiding. By combining the proposed GAN inversion method with traditional steganography, we can easily implement the application of securely hiding an image with $ 1024\times1024 $ resolution to another image.}
	\label{fig:deephidingillustrate}
\end{figure*}

\noindent\textbf{Effectiveness of Our Encoder.} For the non-face domains, we use the Stanford Cars \cite{Krause20133DOR}, LSUN \cite{yu2015lsun} Cat, and LSUN \cite{yu2015lsun} Horse datasets to evaluate the performance of the different methods, with the number of training sets being $8144$, $10000$, and $10000$, respectively, and the number of test sets all being $2,000$. To compare the performance of different encoder network structures, we retrain pSp \cite{Richardson2020EncodingIS}, IDGI \cite{Zhu2020InDomainGI}, and our encoder on these datasets. These methods all use common ($ \ell_2 $, LPIPS) losses as training constraints and the training strategies for these methods are completely identical, which are consistent with the strategy we use for the face domain. Note that for each non-face domain we use the official StyleGAN2 as the generator, which is trained by images from that domain.

\begin{figure}[t]
	\begin{center}
		\setlength{\tabcolsep}{0.5pt}
		\begin{tabular}{m{0.3cm}<{\centering}m{1.58cm}<{\centering}m{1.58cm}<{\centering}m{1.58cm}<{\centering}m{1.58cm}<{\centering}m{1.58cm}<{\centering}}
			& \footnotesize{Carrier C} & \footnotesize{Stego S} & \footnotesize{$S-C$} & \footnotesize{Secret} & \footnotesize{Recovered}
			\\
			%			\small{UDH \cite{Zhang2020UDHUD}}
			\raisebox{0.4cm}{\rotatebox[origin=c]{90}{\footnotesize{{UDH \cite{Zhang2020UDHUD}}}}}
			&\includegraphics[width=1.54cm]{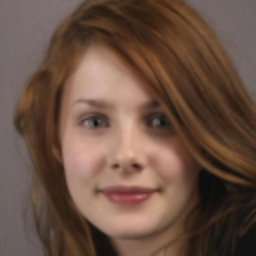}
			&\includegraphics[width=1.54cm]{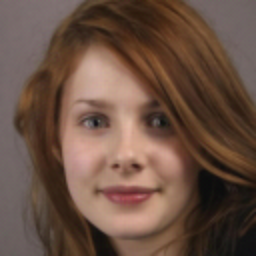}
			&\includegraphics[width=1.54cm]{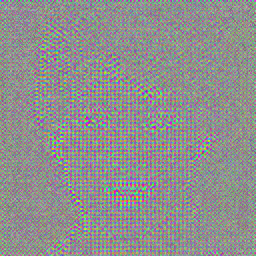}
			&\includegraphics[width=1.54cm]{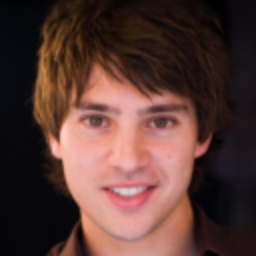}
			&\includegraphics[width=1.54cm]{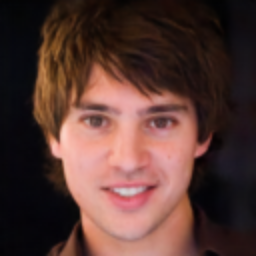}
			\\
			%			\small{DDH \cite{Baluja2017HidingII}}
			\raisebox{0.4cm}{\rotatebox[origin=c]{90}{\footnotesize{{DDH \cite{Baluja2017HidingII}}}}}
			&\includegraphics[width=1.54cm]{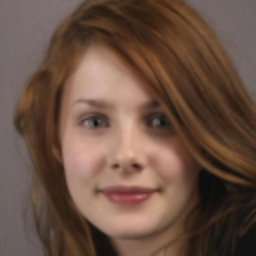}
			&\includegraphics[width=1.54cm]{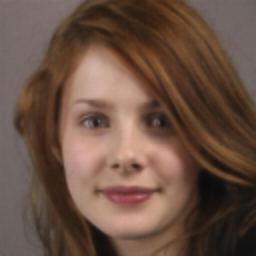}
			&\includegraphics[width=1.54cm]{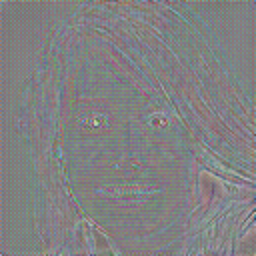}
			&\includegraphics[width=1.54cm]{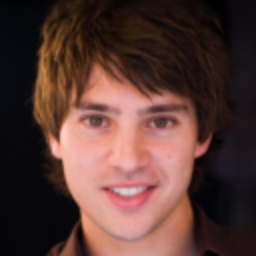}
			&\includegraphics[width=1.54cm]{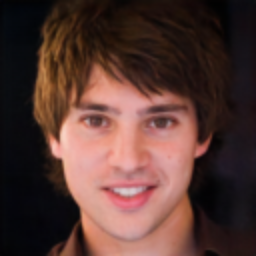}
			\\
			%			\small{Ours}
			\raisebox{0.5cm}{\rotatebox[origin=c]{90}{\footnotesize{{E2Style}}}}
			&\includegraphics[width=1.54cm]{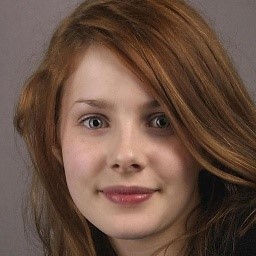}
			&\includegraphics[width=1.54cm]{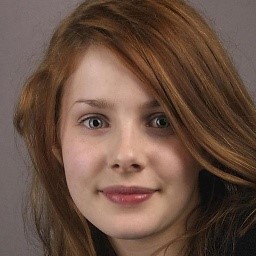}
			&\includegraphics[width=1.54cm]{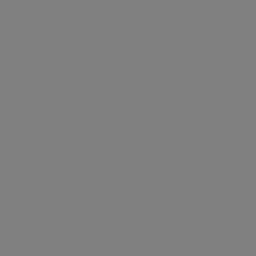}
			&\includegraphics[width=1.54cm]{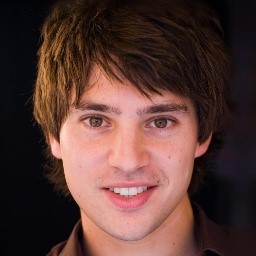}
			&\includegraphics[width=1.54cm]{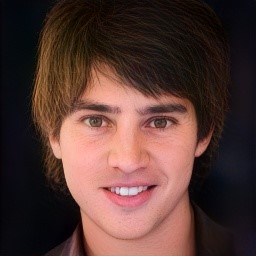}
			\\
			
		\end{tabular}
	\end{center}
	
	\caption{Qualitative comparison of deep hiding. Stego is obtained by hiding Secret into the Carrier. Please zoom in to get a better view of the residuals between the stego image $S$ and the original carrier image $C$.} 
	
	\label{fig:deephiding}
	
\end{figure}

Quantitative and qualitative comparisons are shown in Table \ref{tab:otherdataset} and Figure \ref{fig:otherdatasetfigure}. Obviously, doing inversion tasks on these non-face domains is more challenging due to the more complex and diverse datasets. However, this does not hinder us from comparing the performance of different network structure designs. Thanks to our improvements in network structure, our encoder has the best performance and the lowest network parameters of all forward-based methods.

\vspace{0.5em}

\noindent\textbf{Extensibility of Multi-Layer Face Parsing Loss.} The essence of the multi-layer face parsing loss is showing that local supervision is useful for GAN inversion. By replacing the multi-layer face parsing loss into multi-layer segmentation loss with a pretrained segmentation model \cite{DBLP:conf/bmvc/NekrasovS018}, we find the perceptual quality of inversion results on these non-face datasets can also be significantly improved, although there is a slight decrease in PSNR, SSIM. As shown in Figure \ref{fig:seg-ablation}, the introduction of multi-layer segmentation loss enables the network to pay more attention to salient objects that are of more interest to humans, resulting in a large improvement in perceptual quality.

\section{Applications}

GAN inversion provides the community with new ideas and perspectives for solving image editing problems. With our fast and accurate GAN inversion technique as the foundation, many downstream image processing tasks will be benefited greatly. 
In this section, we show some interesting and practical applications to demonstrate the potentials of our E2Style, including: secure deep hiding, image manipulation, image restoration, and image translation. Except image translation tasks, all applications use FFHQ \cite{Karras2019ASG} as the training dataset and use CelebA-HQ \cite{Karras2018ProgressiveGO} as the test dataset.

\subsection{Secure Deep Hiding}

\begin{algorithm}[h]
	\SetAlgoLined
	\KwIn{secret image $I_s$; carrier image $I_c$; a pretrained StyleGAN2 generator $ G(\cdot) $; our encoder $ E(\cdot) $; traditional steganography embedding and extraction algorithms $ Emb(\cdot, \cdot), Ext(\cdot)$.}
	\KwOut{reconstructed secret image $I_r$.}
	$W_{s}=E(I_s)$\;
	$W_{as}=\lfloor W_{s} \rfloor$\;
	$I_{cs}=Emb(I_c, W_{as})$\;
	The sender sends $I_{cs}$, then the recipient receives $I_{cs}$\;
	$W_{as}=Ext(I_{cs})$\;
	$I_r=G(W_{as})$.
	\caption{Secure Deep Hiding}
	\label{alg:secure_deep_hiding}
\end{algorithm}

In this application, the goal is to hide one secret image into one carrier image in an imperceptible way. 
In contrast to traditional steganography \cite{Holub2012DesigningSD,Holub2014UniversalDF,Sedighi2016ContentAdaptiveSB}, which embeds secret messages with small capacity, deep hiding aims at large capacity hidden messages, such as hiding a secret image with $ 1024\times1024 $ resolution into a carrier image.
Existing deep hiding schemes \cite{Baluja2017HidingII,Zhang2020UDHUD,zhang2020model} either directly concatenate two images together and feed them into the network, or feed only the secret image into the network for the carrier-agnostic purpose, which have weak imperceptibility and undetectability, therefore the security cannot be guaranteed. Empowered by our excellent inversion quality, we can combine the proposed method with traditional steganography to propose a novel application: secure deep hiding, which overcomes the aforementioned drawbacks. Our proposed secure deep hiding framework is shown in Figure \ref{fig:deephidingillustrate}. \wty{Alg.~\ref{alg:secure_deep_hiding} shows the details of the algorithm.} Specifically, the whole hiding process is divided into four steps:  
%1) The sender uses the proposed method to obtain the latent code $ W_{s} $ of the secret image; 2) The sender hides $ W_{s} $ into the carrier image using the traditional steganography method \cite{Filler2011MinimizingAD} to get the stego image, and then sends the stego image out; 3)After receiving the stego image, the recipient recovers $ W_{s} $ accurately using the corresponding extraction algorithm \cite{Filler2011MinimizingAD}; 4) The recipient feeds $ W_{s} $ to the same pre-trained StyleGAN for recovering the secret image.
 \begin{itemize}
		\item [1)] The sender selects the secret image to be sent and then uses the proposed GAN inversion method to get its latent code $ W_{s} \in \mathbb{R}^{18\times512} $. In order to reduce the amount of secret message data while not causing a significant visual impact on the reconstruction result, the sender keeps the integer part of $ W_{s} $ to obtain $ W_{as} \in \mathbb{Z}^{18\times512} $.
		\item [2)] The sender selects a carrier image, hides $ W_{as} $ in the carrier image using the traditional steganography method \cite{Filler2011MinimizingAD} to get the stego image $I_{cs}$, and then sends the stego image out.
		\item [3)] After receiving the stego image, the recipient obtains $ W_{as} $ accurately using the corresponding extraction algorithm \cite{Filler2011MinimizingAD}.
		\item [4)] The recipient feeds $ W_{as} $ to the StyleGAN which has the same network weight as the one inverted by the sender to reconstruct the secret image.
\end{itemize}
In this process, the pre-trained StyleGAN indeed acts as the encryption and decryption codebook.

Figure \ref{fig:deephiding} shows the visual comparison of our method with universal deep hiding (UDH) \cite{Zhang2020UDHUD} and dependent deep hiding (DDH) \cite{Baluja2017HidingII}. With the benefit of the proposed high-quality GAN inversion method, our deep hiding scheme achieves comparable reconstruction quality with UDH and DDH. But as shown in the third column of Figure \ref{fig:deephiding}, 
UDH and DDH either expose the secret image or the carrier image, while our scheme has better imperceptibility and undetectability.
More importantly, the traditional steganography we use provides theoretical security, whereas UDH and DDH do not have it because they are both based on deep learning networks. To summarize, by combining the proposed GAN inversion method with traditional steganography, we can easily implement the application of securely hiding an image with $ 1024\times1024 $ resolution to another image, which is referred to as \textit{Secure Deep Hiding}.
\subsection{Image Manipulation}

\begin{figure}[t]
	\begin{center}
		\setlength{\tabcolsep}{0.5pt}
		\begin{tabular}{cccccc}
			\small{Source} & \small{Target} & \small{FSwap} & \small{FShifter} & \small{Pasted} & \small{Ours}
			\\
			\includegraphics[width=1.34cm]{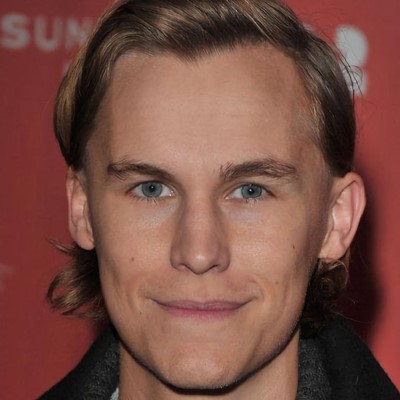}
			&\includegraphics[width=1.34cm]{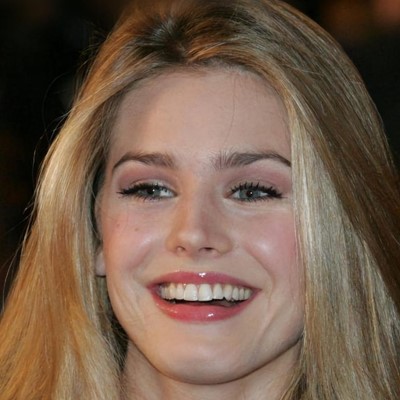}
			&\includegraphics[width=1.34cm]{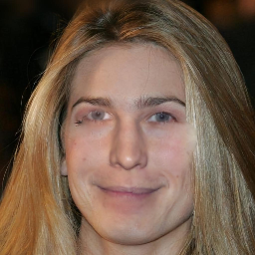}
			&\includegraphics[width=1.34cm]{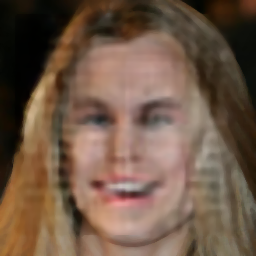}
			&\includegraphics[width=1.34cm]{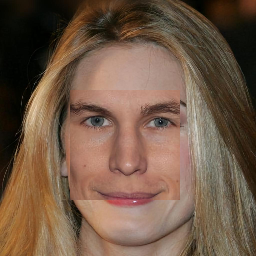}
			&\includegraphics[width=1.34cm]{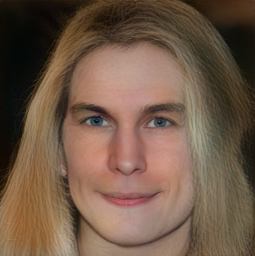}
			\\
			
			\includegraphics[width=1.34cm]{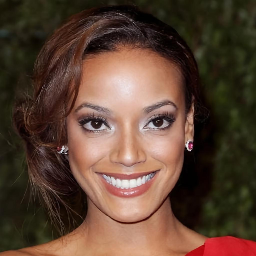}
			&\includegraphics[width=1.34cm]{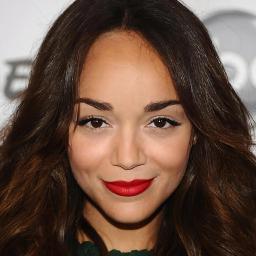}
			&\includegraphics[width=1.34cm]{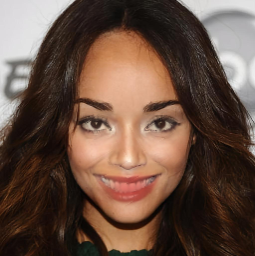}
			&\includegraphics[width=1.34cm]{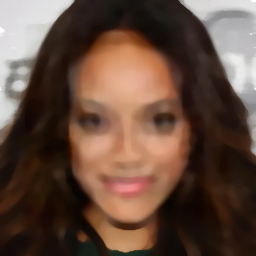}
			&\includegraphics[width=1.34cm]{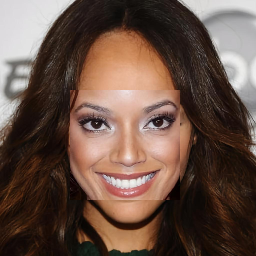}
			&\includegraphics[width=1.34cm]{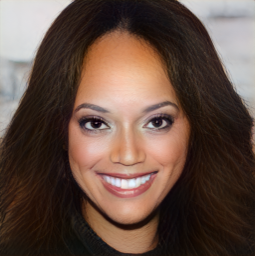}
			\\
			
			\includegraphics[width=1.3cm]{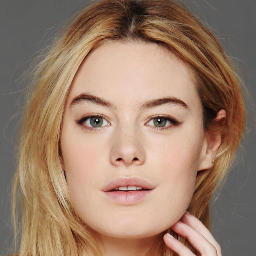}
			&\includegraphics[width=1.3cm]{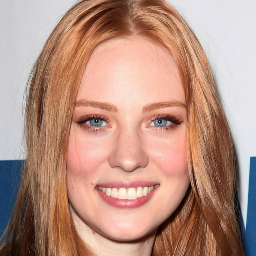}
			&\includegraphics[width=1.3cm]{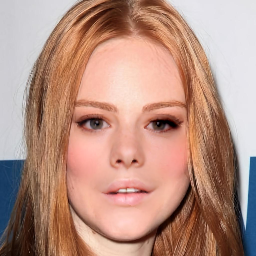}
			&\includegraphics[width=1.3cm]{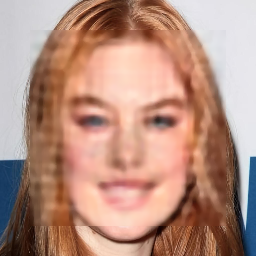}
			&\includegraphics[width=1.3cm]{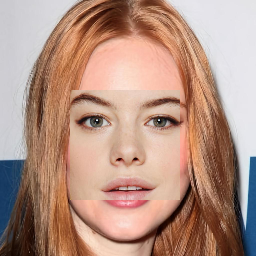}
			&\includegraphics[width=1.3cm]{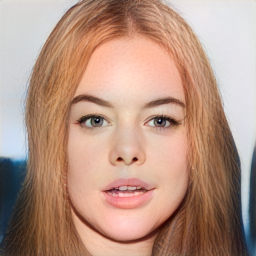}
			\\
		\end{tabular}
	\end{center}
	
	\caption{Comparison of our approach with FSwap \cite{FaceSwap} and FShifter \cite{Li2019FaceShifterTH} on semantic face swapping.} 
	\label{fig:faceswapping}
	
\end{figure}

% \vspace{0.5em}

\begin{algorithm}[h]
	\SetAlgoLined
	\KwIn{source image $I_s$; target image $I_t$; a pretrained StyleGAN2 generator $ G(\cdot) $; our encoder $ E(\cdot) $.}
	\KwOut{identity-swapped face image $I_{is}$.}
	Crop out the center area of $I_s$ and paste it onto $I_t$ to get the pasted image $I_p$\;
	$I_{is}=G(E(I_p))$.
	\caption{Semantic Face Swapping}
	\label{alg:semantic_face_swapping}
\end{algorithm}

\noindent\textbf{Semantic Face Swapping.} Semantic face swapping is intended to replace the identity of the target image with that of the source image, however, it does not guarantee strict alignment of facial attributes such as expression, lighting, head pose, \etc. With the pretrained StyleGAN2 as a strong face reconstruction prior, we find our method can also be used to create more natural face swapping results easily. \wty{Alg.~\ref{alg:semantic_face_swapping} shows the details of the algorithm.} Specifically, we crop out the central area of the source face image and paste it directly into the same position of the target face image to create a pasted image. To eliminate the obvious artifacts present in the pasted image, we use the proposed GAN inversion technique to reconstruct the pasted image. As shown in Figure \ref{fig:faceswapping}, the reconstructed image perfectly blends the pasted face with its surrounding area. Compared to traditional geometric transformation based FaceSwap \cite{FaceSwap} and learning-based FaceShifter \cite{Li2019FaceShifterTH}, our results have higher identity consistency with the source image and better image quality.

\begin{figure}[t]
	\centering
	\includegraphics[width=\columnwidth]{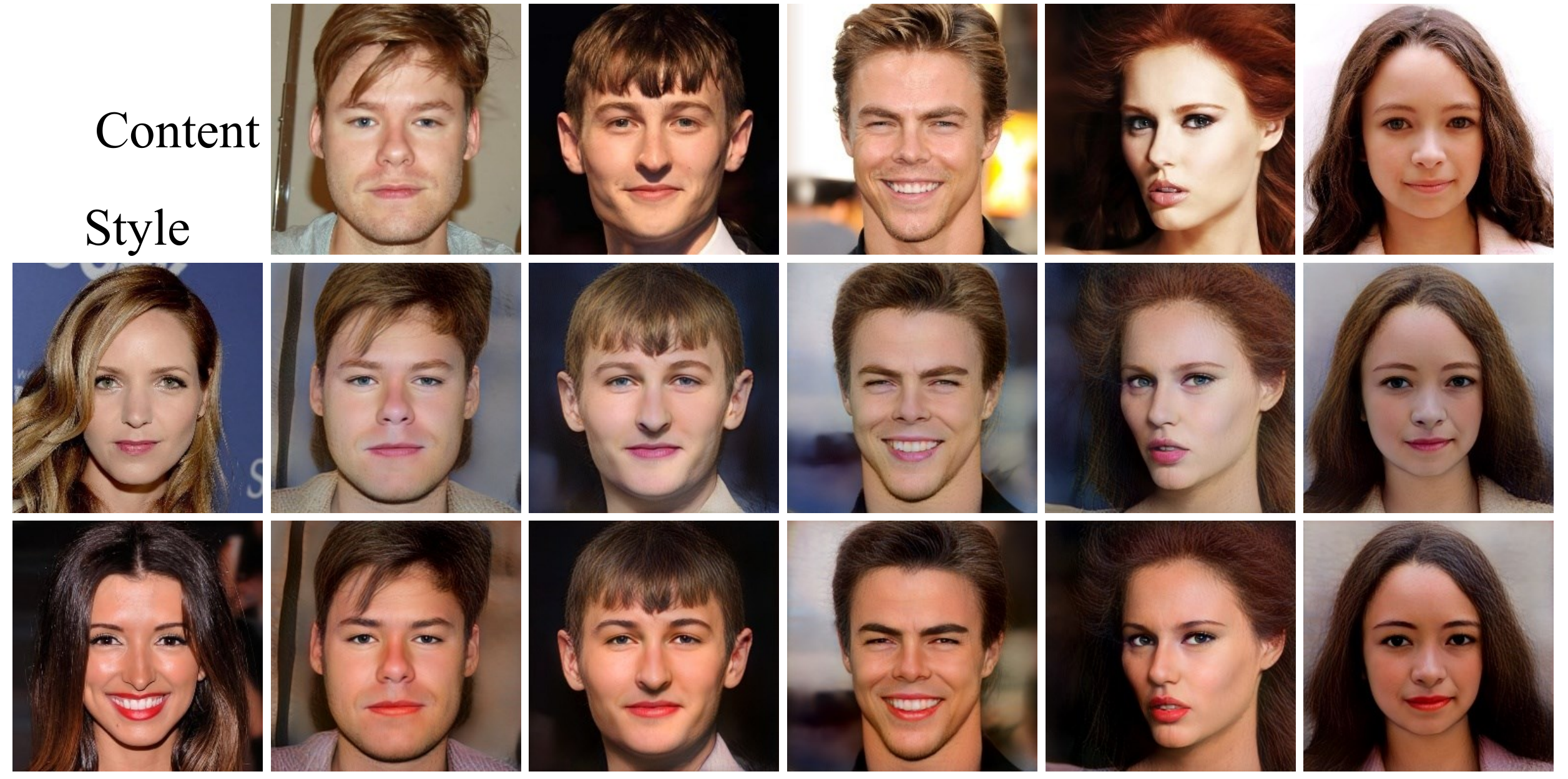} 
	
	\caption{Style mixing results. All the content images are coated with the same color lipstick as the style image.} 
	\label{fig:stylemixing}
	
\end{figure}

% \vspace{0.5em}

\begin{algorithm}[h]
	\SetAlgoLined
	\KwIn{style image $I_s$; content image $I_c$; a pretrained StyleGAN2 generator $ G(\cdot) $; our encoder $ E(\cdot) $; style mixing operation $Mix(\cdot, \cdot)$.}
	\KwOut{style mixing result image $I_{m}$.}
	$I_{m}=G(Mix(E(I_s), E(I_c)))$.
	\caption{Style Mixing}
	\label{alg:style_mixing}
\end{algorithm}

\noindent\textbf{Style Mixing.} Style mixing aims to transfer the low-level appearance characteristics from the style image to the content image. \wty{Alg.~\ref{alg:style_mixing} shows the details of the algorithm.} In detail, after obtaining the respective latent codes of the style image and the content image, we replace the last $ 11 $ layers of the latent codes corresponding to the content image with those of the style image. A specific example is shown in Figure \ref{fig:stylemixing}, where all the content images are coated with the same color lipstick as the style image.

% \vspace{0.5em}

\begin{algorithm}[h]
	\SetAlgoLined
	\KwIn{real image $I_A$; real image $I_B$; a pretrained StyleGAN2 generator $ G(\cdot) $; our encoder $ E(\cdot) $; blending parameter $ \lambda $.}
	\KwOut{face interpolation result image $I_{i}$.}
	\For{$\lambda=0$ to $1$}
	{$I_{i}=G(\lambda E(I_B)+(1-\lambda)E(I_A))$;}
	\caption{Face Interpolation}
	\label{alg:face_interpolation}
\end{algorithm}

\begin{figure*}[t]
	\centering
	\includegraphics[width=\textwidth]{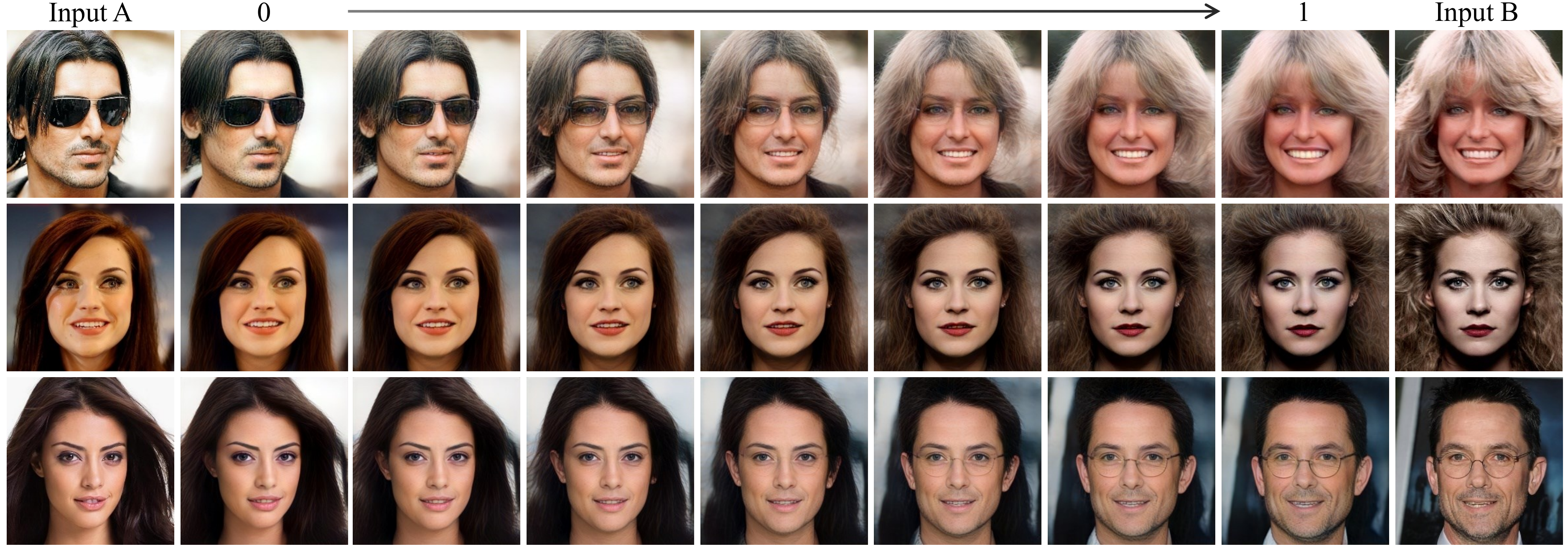} 
	\caption{Face interpolation results. By gradually increasing the blending parameter $ \lambda $ from $ 0 $ to $ 1 $, we achieve the morphing effect from input A to B.} 
	\label{fig:interpolate}
\end{figure*}

\noindent\textbf{Face Interpolation.} Given two real-world images A and B,
 Face interpolation is intended to complete the gradual morphing from input A to input B. \wty{Alg.~\ref{alg:face_interpolation} shows the details of the algorithm.}
 we use the proposed GAN inversion method to obtain their respective latent codes $ W_{A},W_{B} \in \mathcal{W+} $. Then, we combine the two latent codes by linear weighting to generate the intermediate latent code $ W_{I}=\lambda W_{B}+(1-\lambda)W_{A}$. Finally, the image corresponding to the intermediate latent code is generated. By gradually increasing the blending parameter $ \lambda $ from $ 0 $ to $ 1 $, we can achieve the gradual morphing effect from input A to input B, as shown in Figure \ref{fig:interpolate}.

% \vspace{0.5em}

\subsection{Image Restoration}

\begin{algorithm}[h]
	\SetAlgoLined
	\KwIn{impaired image $I_i$; a pretrained StyleGAN2 generator $ G(\cdot) $; our encoder for the specified restoration task $ E_{r}(\cdot) $.}
	\KwOut{restored image $I_{r}$.}
	$I_{r}=G(E_{r}(I_i))$.
	\caption{Image Restoration}
	\label{alg:image_restoration}
\end{algorithm}

By introducing corresponding data augmentations to the input images during the training phase, our E2Style can be easily extended to perform a number of image restoration tasks, such as colorization, inpainting, super resolution. \wty{Alg.~\ref{alg:image_restoration} shows the details of the algorithm in the inference phase.} Here we qualitatively compare our E2Style with pSp \cite{Richardson2020EncodingIS}, e4e \cite{Tov2021DesigningAE}, and Restyle \cite{alaluf2021restyle} on these tasks, both of which use the same data augmentations. Figure \ref{fig:restorationandtranslation} shows the visual comparison of our E2Style with other methods on these three tasks.

\vspace{0.5em}
\noindent\textbf{Colorization.} Image colorization aims to convert the single-channel gray-scale image into a semantically reasonable three-channel color image. For the colorization task, we use the color image as the target and the corresponding grayscale image as input when training the network, so the goal of the encoder is no longer to reconstruct the input image accurately but to find the most appropriate latent code which can colorize the input grayscale image reasonably. Compared with other methods, E2Style accomplishes reasonable coloring for the grayscale image while ensuring consistency in identity and local details (\eg, glasses).

\vspace{0.5em}
\noindent\textbf{Inpainting.}  Given an incomplete image, the task of inpainting is to recover the missing pixels so that the recovered part is compatible with the known pixels. To perform inpainting using the GAN inversion framework, we employ a degraded transformation to the input image, in which a region is randomly selected and the pixel value of that region is set to $ 0 $ during the training process. As illustrated in Figure \ref{fig:restorationandtranslation}, E2Style achieves reasonable filling of unknown regions while keeping existing pixel values unchanged as much as possible.

\vspace{0.5em}
\noindent\textbf{Super Resolution.} The goal of super resolution is to reconstruct the corresponding high-resolution image from the observed low-resolution image. Following the data augmentation method of pSp, we randomly downsample the input image by $ \times1 $, $ \times2 $, $ \times4 $, $ \times8 $, and $ \times16 $ and use the original resolution image as the target during training. Compared to other methods (Figure \ref{fig:restorationandtranslation}), our results are more realistic and with better facial details, \eg, the eyes in the shown example are more faithfully aligned with the low-resolution image, \wty{and the glasses of another example are faithfully reconstructed.}

%\vspace{0.5em}
%
%\noindent\textbf{Denoising.} For the denoising task, we randomly add pretzel noise or Gaussian noise to the input image when training the encoder. Compared to pSp (Figure \ref{fig:restorationandtranslation}), our method does a good job of removing noise from the input image while retaining a large degree of information about the original identity and local details (\eg, mouth).

\subsection{Image Translation}

\begin{algorithm}[h]
	\SetAlgoLined
	\KwIn{sketch or segmentation map $I_s$; a pretrained StyleGAN2 generator $ G(\cdot) $; our encoder for the specified image translation task $ E_{t}(\cdot) $.}
	\KwOut{translated image $I_{t}$.}
	$I_{t}=G(E_{t}(I_s))$.
	\caption{Image Translation}
	\label{alg:image_translation}
\end{algorithm}

\begin{figure*}[t]
	\begin{center}
		\setlength{\tabcolsep}{0.5pt}
		\begin{tabular}{m{0.3cm}<{\centering}m{1.7cm}<{\centering}m{1.7cm}<{\centering}m{1.7cm}<{\centering}m{1.7cm}<{\centering}m{1.7cm}<{\centering}m{1.7cm}<{\centering}m{1.7cm}<{\centering}m{1.7cm}<{\centering}m{1.7cm}<{\centering}m{1.7cm}<{\centering}}
			
		& \multicolumn{2}{c}{\footnotesize{Colorization}} & \multicolumn{2}{c}{\footnotesize{Inpainting}} & \multicolumn{2}{c}{\footnotesize{Super Resolution}} & \multicolumn{4}{c}{\footnotesize{Image Translation}}
		\\		
			\raisebox{0.2cm}{\rotatebox[origin=c]{90}{\footnotesize{{Input}}}}
			&\includegraphics[width=1.68cm]{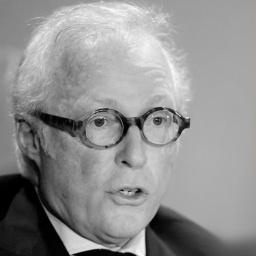}
			&\includegraphics[width=1.68cm]{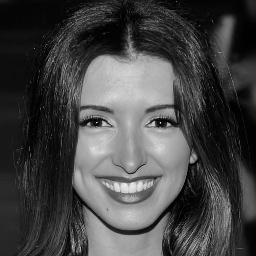}
			&\includegraphics[width=1.68cm]{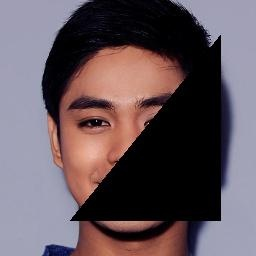}
			&\includegraphics[width=1.68cm]{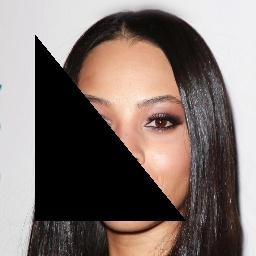}
			&\includegraphics[width=1.68cm]{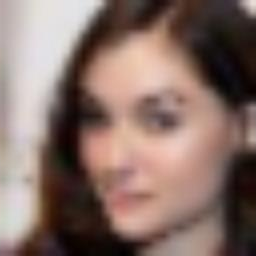}
			&\includegraphics[width=1.68cm]{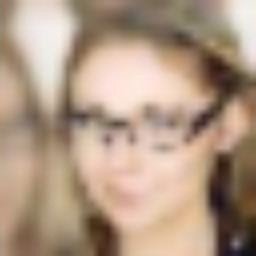}
			&\includegraphics[width=1.68cm]{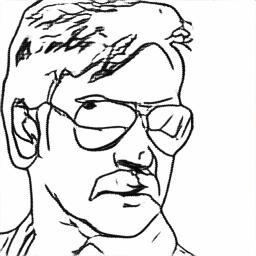}
			&\includegraphics[width=1.68cm]{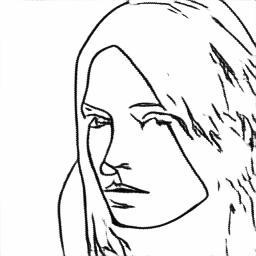}
			&\includegraphics[width=1.68cm]{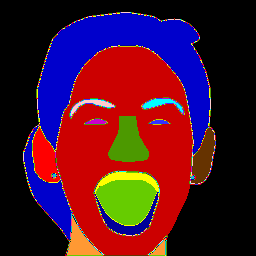}
			&\includegraphics[width=1.68cm]{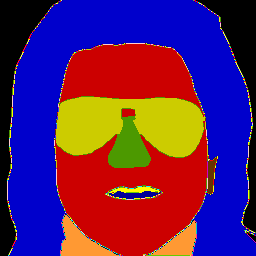}
			\\
			
			\raisebox{0.6cm}{\rotatebox[origin=c]{90}{\footnotesize{{Restyle \cite{alaluf2021restyle}}}}}
			&\includegraphics[width=1.68cm]{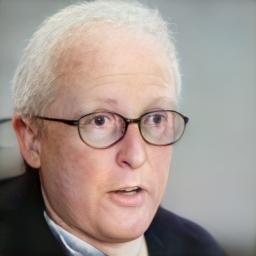}
			&\includegraphics[width=1.68cm]{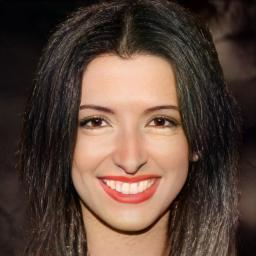}
			&\includegraphics[width=1.68cm]{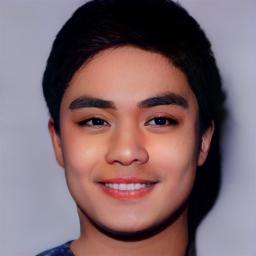}
			&\includegraphics[width=1.68cm]{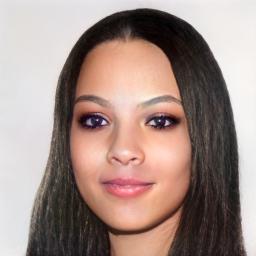}
			&\includegraphics[width=1.68cm]{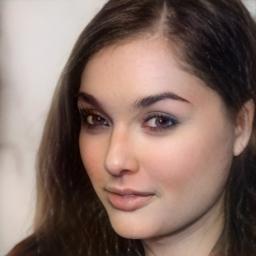}
			&\includegraphics[width=1.68cm]{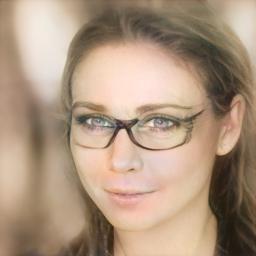}
			&\includegraphics[width=1.68cm]{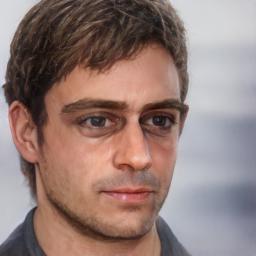}
			&\includegraphics[width=1.68cm]{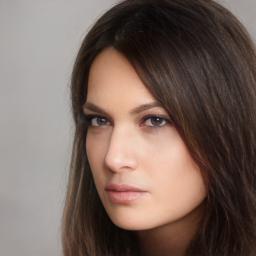}
			&\includegraphics[width=1.68cm]{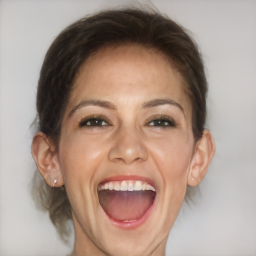}
			&\includegraphics[width=1.68cm]{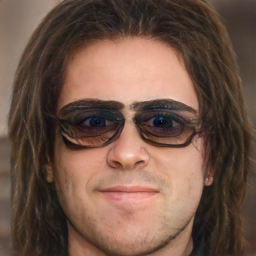}
			\\			

			\raisebox{0.39cm}{\rotatebox[origin=c]{90}{\footnotesize{{e4e \cite{Tov2021DesigningAE}}}}}
			&\includegraphics[width=1.68cm]{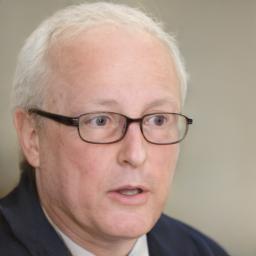}
			&\includegraphics[width=1.68cm]{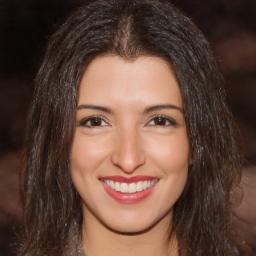}
			&\includegraphics[width=1.68cm]{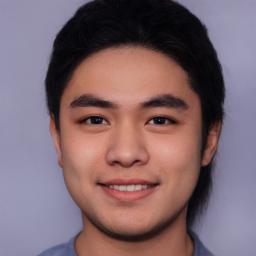}
			&\includegraphics[width=1.68cm]{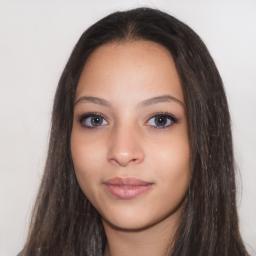}
			&\includegraphics[width=1.68cm]{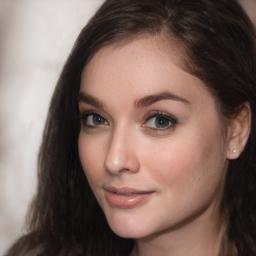}
			&\includegraphics[width=1.68cm]{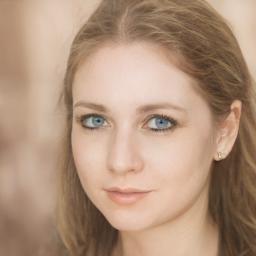}
			&\includegraphics[width=1.68cm]{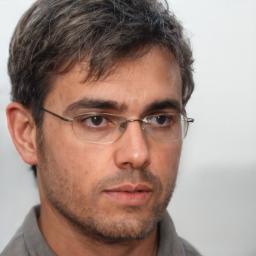}
			&\includegraphics[width=1.68cm]{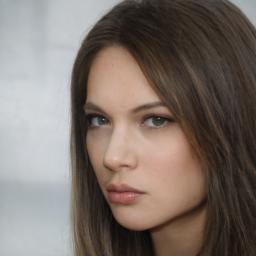}
			&\includegraphics[width=1.68cm]{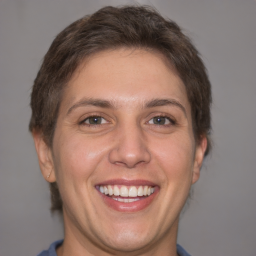}
			&\includegraphics[width=1.68cm]{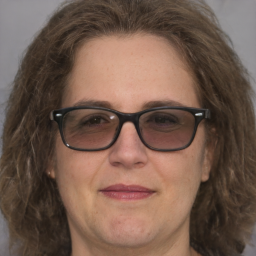}
			\\
			
			\raisebox{0.45cm}{\rotatebox[origin=c]{90}{\footnotesize{{pSp \cite{Richardson2020EncodingIS}}}}}
			&\includegraphics[width=1.68cm]{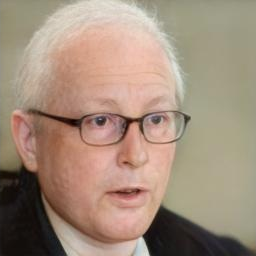}
			&\includegraphics[width=1.68cm]{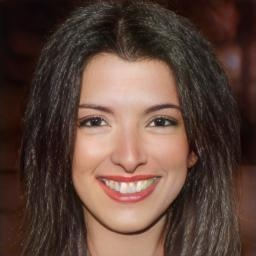}
			&\includegraphics[width=1.68cm]{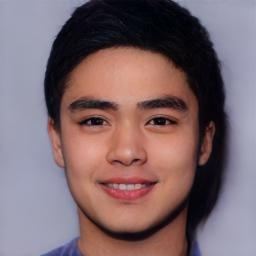}
			&\includegraphics[width=1.68cm]{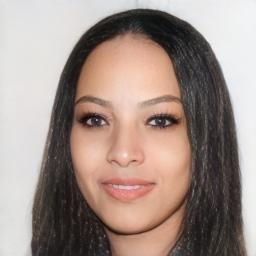}
			&\includegraphics[width=1.68cm]{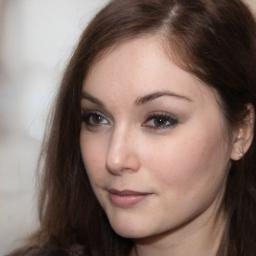}
			&\includegraphics[width=1.68cm]{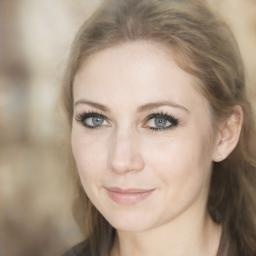}
			&\includegraphics[width=1.68cm]{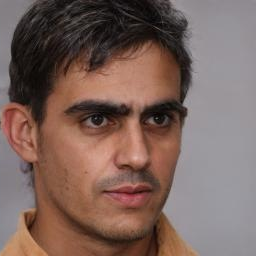}
			&\includegraphics[width=1.68cm]{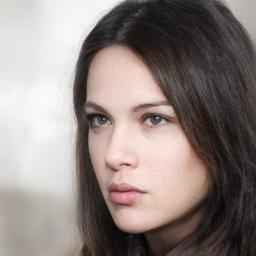}
			&\includegraphics[width=1.68cm]{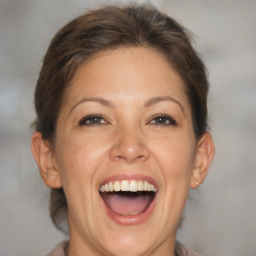}
			&\includegraphics[width=1.68cm]{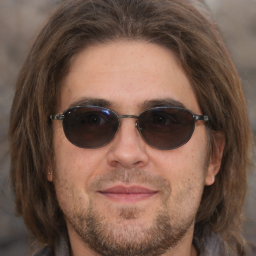}
			\\
			
			\raisebox{0.5cm}{\rotatebox[origin=c]{90}{\footnotesize{{E2Style}}}}
			&\includegraphics[width=1.68cm]{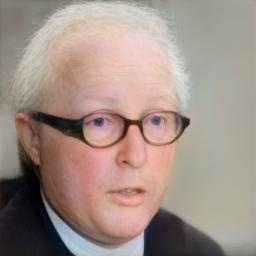}
			&\includegraphics[width=1.68cm]{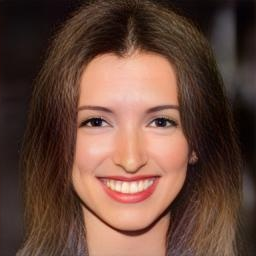}
			&\includegraphics[width=1.68cm]{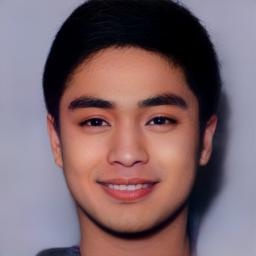}
			&\includegraphics[width=1.68cm]{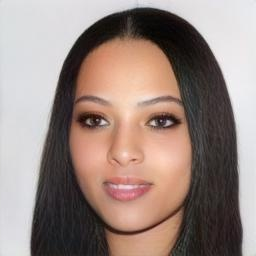}
			&\includegraphics[width=1.68cm]{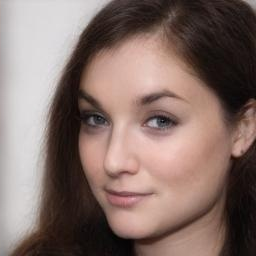}
			&\includegraphics[width=1.68cm]{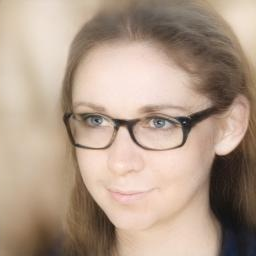}
			&\includegraphics[width=1.68cm]{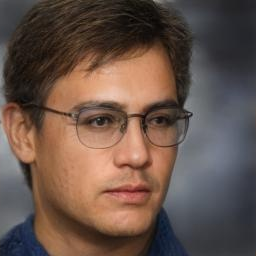}
			&\includegraphics[width=1.68cm]{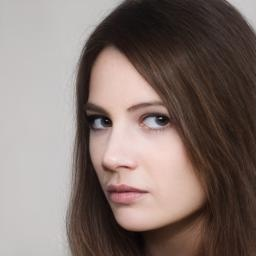}
			&\includegraphics[width=1.68cm]{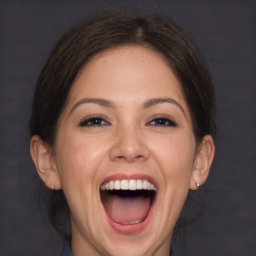}
			&\includegraphics[width=1.68cm]{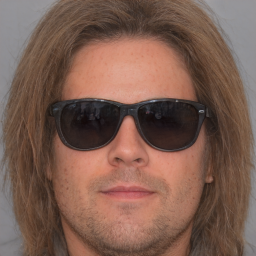}
			\\
		\end{tabular}
	\end{center}
	\caption{\wty{Comparison of our E2Style with pSp \cite{Richardson2020EncodingIS}, e4e \cite{Tov2021DesigningAE}, and Restyle \cite{alaluf2021restyle} on image restoration including colorization, inpainting, super resolution and image translation tasks.}} 
	\label{fig:restorationandtranslation}
\end{figure*}

By replacing the input with the corresponding sketch or segmentation label map of the image during the network training phase, our framework can also perform image translation tasks. Given that there is no relevant dataset for the FFHQ, we perform this task on the CelebA-HQ, of which $3,000$ images are reserved for testing purposes. Specifically, we follow  \cite{Richardson2020EncodingIS} to generate the sketch dataset of CelebA-HQ, while the segmentation label maps are from the CelebAMask-HQ dataset \cite{CelebAMask-HQ}. Regarding the training loss, it is the same as the GAN inversion task except that the multi-layer identity loss is removed. \wty{Alg.~\ref{alg:image_translation} shows the details of the algorithm in the inference phase.} The visual comparison results are shown in Figure \ref{fig:restorationandtranslation}. Compared to other methods, it is clear that our results are more faithfully aligned to the respective semantics of the input, \eg the glasses in the sketch-to-image example.

\section{Conclusion}
In this paper, we propose E2Style, which improves GAN inversion in terms of efficiency and effectiveness. Such improvements come from three aspects: 1) designing a more efficient GAN inversion network with a shallow backbone, hierarchical latent code regression, and efficient prediction heads; 2) introducing multi-layer identity loss and multi-layer parsing loss; and 3) purely feed-forward-based multi-stage refinement. Extensive evaluation and applications demonstrate that our E2Style performs much better than existing feed-forward-based methods and comparably to state-of-the-art optimization-based methods with higher efficiency. \aq{In the future, more emphasis should be put on creating a universal approach to encode all kinds of images into the latent space of a StyleGAN model pre-trained in the corresponding domain.}
% \aq{In the future, we are interested in developing an encoder for multi-modal latent spaces.}
%simple

%\appendices
%\section{}
%\section*{Acknowledgment}

\ifCLASSOPTIONcaptionsoff
  \newpage
\fi

\bibliographystyle{IEEEtran}
\bibliography{IEEEabrv,egbib}

\begin{thebibliography}{10}
\providecommand{\url}[1]{#1}
\csname url@samestyle\endcsname
\providecommand{\newblock}{\relax}
\providecommand{\bibinfo}[2]{#2}
\providecommand{\BIBentrySTDinterwordspacing}{\spaceskip=0pt\relax}
\providecommand{\BIBentryALTinterwordstretchfactor}{4}
\providecommand{\BIBentryALTinterwordspacing}{\spaceskip=\fontdimen2\font plus
\BIBentryALTinterwordstretchfactor\fontdimen3\font minus
  \fontdimen4\font\relax}
\providecommand{\BIBforeignlanguage}[2]{{%
\expandafter\ifx\csname l@#1\endcsname\relax
\typeout{** WARNING: IEEEtran.bst: No hyphenation pattern has been}%
\typeout{** loaded for the language `#1'. Using the pattern for}%
\typeout{** the default language instead.}%
\else
\language=\csname l@#1\endcsname
\fi
#2}}
\providecommand{\BIBdecl}{\relax}
\BIBdecl

\bibitem{Karras2019ASG}
T.~Karras, S.~Laine, and T.~Aila, ``A style-based generator architecture for
  generative adversarial networks,'' \emph{IEEE/CVF Conference on Computer
  Vision and Pattern Recognition (CVPR)}, 2019.

\bibitem{Karras2020AnalyzingAI}
T.~Karras, S.~Laine, M.~Aittala, J.~Hellsten, J.~Lehtinen, and T.~Aila,
  ``Analyzing and improving the image quality of stylegan,'' \emph{2020
  IEEE/CVF Conference on Computer Vision and Pattern Recognition (CVPR)}, pp.
  8107--8116, 2020.

\bibitem{Shen2020InterpretingTL}
Y.~Shen, J.~Gu, X.~Tang, and B.~Zhou, ``Interpreting the latent space of gans
  for semantic face editing,'' \emph{2020 IEEE/CVF Conference on Computer
  Vision and Pattern Recognition (CVPR)}, pp. 9240--9249, 2020.

\bibitem{Abdal2019Image2StyleGANHT}
R.~Abdal, Y.~Qin, and P.~Wonka, ``Image2stylegan: How to embed images into the
  stylegan latent space?'' \emph{2019 IEEE/CVF International Conference on
  Computer Vision (ICCV)}, pp. 4431--4440, 2019.

\bibitem{Abdal2020Image2StyleGANHT}
------, ``Image2stylegan++: How to edit the embedded images?'' \emph{2020
  IEEE/CVF Conference on Computer Vision and Pattern Recognition (CVPR)}, pp.
  8293--8302, 2020.

\bibitem{Nitzan2020FaceID}
Y.~Nitzan, A.~Bermano, Y.~Li, and D.~Cohen-Or, ``Face identity disentanglement
  via latent space mapping,'' \emph{ACM Transactions on Graphics (TOG)},
  vol.~39, pp. 1 -- 14, 2020.

\bibitem{Guan2020CollaborativeLF}
S.~Guan, Y.~Tai, B.~Ni, F.~Zhu, F.~Huang, and X.~Yang, ``Collaborative learning
  for faster stylegan embedding,'' \emph{ArXiv}, vol. abs/2007.01758, 2020.

\bibitem{Yang2021ADN}
N.~Yang, Z.~Zheng, M.~Zhou, X.~Guo, L.~Qi, and T.~Wang, ``A domain-guided
  noise-optimization-based inversion method for facial image manipulation,''
  \emph{IEEE Transactions on Image Processing}, vol.~30, pp. 6198--6211, 2021.

\bibitem{Pan2020ExploitingDG}
X.~Pan, X.~Zhan, B.~Dai, D.~Lin, C.~C. Loy, and P.~Luo, ``Exploiting deep
  generative prior for versatile image restoration and manipulation,''
  \emph{ArXiv}, vol. abs/2003.13659, 2020.

\bibitem{Richardson2020EncodingIS}
E.~Richardson, Y.~Alaluf, O.~Patashnik, Y.~Nitzan, Y.~Azar, S.~Shapiro, and
  D.~Cohen-Or, ``Encoding in style: A stylegan encoder for image-to-image
  translation,'' in \emph{Proceedings of the IEEE/CVF Conference on Computer
  Vision and Pattern Recognition (CVPR)}, June 2021, pp. 2287--2296.

\bibitem{Zhu2020InDomainGI}
J.~Zhu, Y.~Shen, D.~li~Zhao, and B.~Zhou, ``In-domain gan inversion for real
  image editing,'' in \emph{ECCV}, 2020.

\bibitem{Bau2019SeeingWA}
D.~Bau, J.-Y. Zhu, J.~Wulff, W.~Peebles, H.~Strobelt, B.~Zhou, and A.~Torralba,
  ``Seeing what a gan cannot generate,'' \emph{2019 IEEE/CVF International
  Conference on Computer Vision (ICCV)}, pp. 4501--4510, 2019.

\bibitem{Bau2019SemanticPM}
D.~Bau, H.~Strobelt, W.~Peebles, J.~Wulff, B.~Zhou, J.-Y. Zhu, and A.~Torralba,
  ``Semantic photo manipulation with a generative image prior,'' \emph{ACM
  Transactions on Graphics (TOG)}, vol.~38, pp. 1 -- 11, 2019.

\bibitem{Goodfellow2014GenerativeAN}
I.~J. Goodfellow, J.~Pouget-Abadie, M.~Mirza, B.~Xu, D.~Warde-Farley, S.~Ozair,
  A.~C. Courville, and Y.~Bengio, ``Generative adversarial nets,'' in
  \emph{NIPS}, 2014.

\bibitem{Tran2021OnDA}
N.-T. Tran, V.-H. Tran, N.-B. Nguyen, T.-K. Nguyen, and N.-M. Cheung, ``On data
  augmentation for gan training,'' \emph{IEEE Transactions on Image
  Processing}, vol.~30, pp. 1882--1897, 2021.

\bibitem{Ansari2020ACF}
A.~F. Ansari, J.~Scarlett, and H.~Soh, ``A characteristic function approach to
  deep implicit generative modeling,'' \emph{2020 IEEE/CVF Conference on
  Computer Vision and Pattern Recognition (CVPR)}, pp. 7476--7484, 2020.

\bibitem{Arjovsky2017WassersteinGA}
M.~Arjovsky, S.~Chintala, and L.~Bottou, ``Wasserstein generative adversarial
  networks,'' in \emph{ICML}, 2017.

\bibitem{mao2017least}
X.~Mao, Q.~Li, H.~Xie, R.~Y. Lau, Z.~Wang, and S.~Paul~Smolley, ``Least squares
  generative adversarial networks,'' in \emph{Proceedings of the IEEE
  international conference on computer vision}, 2017, pp. 2794--2802.

\bibitem{Miyato2018SpectralNF}
T.~Miyato, T.~Kataoka, M.~Koyama, and Y.~Yoshida, ``Spectral normalization for
  generative adversarial networks,'' \emph{ArXiv}, vol. abs/1802.05957, 2018.

\bibitem{Brock2019LargeSG}
A.~Brock, J.~Donahue, and K.~Simonyan, ``Large scale gan training for high
  fidelity natural image synthesis,'' \emph{ArXiv}, vol. abs/1809.11096, 2019.

\bibitem{qin2020does}
Y.~Qin, N.~Mitra, and P.~Wonka, ``How does lipschitz regularization influence
  gan training?'' in \emph{European Conference on Computer Vision}.\hskip 1em
  plus 0.5em minus 0.4em\relax Springer, 2020, pp. 310--326.

\bibitem{Schnfeld2020AUB}
E.~Sch{\"o}nfeld, B.~Schiele, and A.~Khoreva, ``A u-net based discriminator for
  generative adversarial networks,'' \emph{2020 IEEE/CVF Conference on Computer
  Vision and Pattern Recognition (CVPR)}, pp. 8204--8213, 2020.

\bibitem{Gulrajani2017ImprovedTO}
I.~Gulrajani, F.~Ahmed, M.~Arjovsky, V.~Dumoulin, and A.~C. Courville,
  ``Improved training of wasserstein gans,'' in \emph{NIPS}, 2017.

\bibitem{wan2020bringing}
Z.~Wan, B.~Zhang, D.~Chen, P.~Zhang, D.~Chen, J.~Liao, and F.~Wen, ``Bringing
  old photos back to life,'' in \emph{Proceedings of the IEEE/CVF Conference on
  Computer Vision and Pattern Recognition}, 2020, pp. 2747--2757.

\bibitem{Tan2019ImprovedAF}
W.~R. Tan, C.~S. Chan, H.~E. Aguirre, and K.~Tanaka, ``Improved artgan for
  conditional synthesis of natural image and artwork,'' \emph{IEEE Transactions
  on Image Processing}, vol.~28, pp. 394--409, 2019.

\bibitem{Hsu2019SiGANSG}
C.-C. Hsu, C.-W. Lin, W.-T. Su, and G.~Cheung, ``Sigan: Siamese generative
  adversarial network for identity-preserving face hallucination,'' \emph{IEEE
  Transactions on Image Processing}, vol.~28, pp. 6225--6236, 2019.

\bibitem{Xu2020MEFGANMI}
H.~Xu, J.~Ma, and X.~Zhang, ``Mef-gan: Multi-exposure image fusion via
  generative adversarial networks,'' \emph{IEEE Transactions on Image
  Processing}, vol.~29, pp. 7203--7216, 2020.

\bibitem{Chen2019GatedGANAG}
X.~Chen, C.~Xu, X.~Yang, L.~Song, and D.~Tao, ``Gated-gan: Adversarial gated
  networks for multi-collection style transfer,'' \emph{IEEE Transactions on
  Image Processing}, vol.~28, pp. 546--560, 2019.

\bibitem{Gao2020RPDGANLT}
X.~Gao, Y.~jie Tian, and Z.~Qi, ``Rpd-gan: Learning to draw realistic paintings
  with generative adversarial network,'' \emph{IEEE Transactions on Image
  Processing}, vol.~29, pp. 8706--8720, 2020.

\bibitem{Lucas2019GenerativeAN}
A.~Lucas, S.~Lx00F3pez-Tapia, R.~Molina, and A.~Katsaggelos, ``Generative
  adversarial networks and perceptual losses for video super-resolution,''
  \emph{IEEE Transactions on Image Processing}, vol.~28, pp. 3312--3327, 2019.

\bibitem{liu2021fiss}
K.~Liu, Z.~Ye, H.~Guo, D.~Cao, L.~Chen, and F.-Y. Wang, ``Fiss gan: A
  generative adversarial network for foggy image semantic segmentation,''
  \emph{IEEE/CAA Journal of Automatica Sinica}, vol.~8, no.~8, pp. 1428--1439,
  2021.

\bibitem{zhang2020mu}
K.~Zhang, Y.~Su, X.~Guo, L.~Qi, and Z.~Zhao, ``Mu-gan: Facial attribute editing
  based on multi-attention mechanism,'' \emph{IEEE/CAA Journal of Automatica
  Sinica}, vol.~8, no.~9, pp. 1614--1626, 2020.

\bibitem{li2021predicting}
J.~Li, Y.~Tao, and T.~Cai, ``Predicting lung cancers using epidemiological
  data: A generative-discriminative framework,'' \emph{IEEE/CAA Journal of
  Automatica Sinica}, vol.~8, no.~5, pp. 1067--1078, 2021.

\bibitem{tan2020michigan}
Z.~Tan, M.~Chai, D.~Chen, J.~Liao, Q.~Chu, L.~Yuan, S.~Tulyakov, and N.~Yu,
  ``Michigan: multi-input-conditioned hair image generation for portrait
  editing,'' \emph{ACM Transactions on Graphics (TOG)}, vol.~39, no.~4, pp.
  95--1, 2020.

\bibitem{tan2021efficient}
Z.~Tan, D.~Chen, Q.~Chu, M.~Chai, J.~Liao, M.~He, L.~Yuan, G.~Hua, and N.~Yu,
  ``Efficient semantic image synthesis via class-adaptive normalization,''
  \emph{IEEE Transactions on Pattern Analysis and Machine Intelligence}, 2021.

\bibitem{tan2021diverse}
Z.~Tan, M.~Chai, D.~Chen, J.~Liao, Q.~Chu, B.~Liu, G.~Hua, and N.~Yu, ``Diverse
  semantic image synthesis via probability distribution modeling,'' in
  \emph{Proceedings of the IEEE/CVF Conference on Computer Vision and Pattern
  Recognition}, 2021.

\bibitem{Deng2009ImageNetAL}
J.~Deng, W.~Dong, R.~Socher, L.~Li, K.~Li, and L.~Fei-Fei, ``Imagenet: A
  large-scale hierarchical image database,'' in \emph{CVPR}, 2009.

\bibitem{Karras2018ProgressiveGO}
T.~Karras, T.~Aila, S.~Laine, and J.~Lehtinen, ``Progressive growing of gans
  for improved quality, stability, and variation,'' \emph{ArXiv}, vol.
  abs/1710.10196, 2018.

\bibitem{Shen2020ClosedFormFO}
Y.~Shen and B.~Zhou, ``Closed-form factorization of latent semantics in gans,''
  \emph{ArXiv}, vol. abs/2007.06600, 2020.

\bibitem{Hrknen2020GANSpaceDI}
E.~H{\"a}rk{\"o}nen, A.~Hertzmann, J.~Lehtinen, and S.~Paris, ``Ganspace:
  Discovering interpretable gan controls,'' \emph{ArXiv}, vol. abs/2004.02546,
  2020.

\bibitem{wang2021cross}
C.~Wang, M.~Chai, M.~He, D.~Chen, and J.~Liao, ``Cross-domain and disentangled
  face manipulation with 3d guidance,'' \emph{IEEE Transactions on
  Visualization and Computer Graphics}, 2021.

\bibitem{Gu2020ImagePU}
J.~Gu, Y.~Shen, and B.~Zhou, ``Image processing using multi-code gan prior,''
  \emph{2020 IEEE/CVF Conference on Computer Vision and Pattern Recognition
  (CVPR)}, pp. 3009--3018, 2020.

\bibitem{Bartz2020OneMT}
C.~Bartz, J.~Bethge, H.~Yang, and C.~Meinel, ``One model to reconstruct them
  all: A novel way to use the stochastic noise in stylegan,'' \emph{ArXiv},
  vol. abs/2010.11113, 2020.

\bibitem{yang2021inversion}
N.~Yang, M.~Zhou, B.~Xia, X.~Guo, and L.~Qi, ``Inversion based on a detached
  dual-channel domain method for stylegan2 embedding,'' \emph{IEEE Signal
  Processing Letters}, vol.~28, pp. 553--557, 2021.

\bibitem{alaluf2021restyle}
Y.~Alaluf, O.~Patashnik, and D.~Cohen-Or, ``Restyle: A residual-based stylegan
  encoder via iterative refinement,'' in \emph{Proceedings of the IEEE/CVF
  International Conference on Computer Vision (ICCV)}, October 2021.

\bibitem{lin2017feature}
T.-Y. Lin, P.~Doll{\'a}r, R.~Girshick, K.~He, B.~Hariharan, and S.~Belongie,
  ``Feature pyramid networks for object detection,'' in \emph{Proceedings of
  the IEEE conference on computer vision and pattern recognition}, 2017, pp.
  2117--2125.

\bibitem{szegedy2015going}
C.~Szegedy, W.~Liu, Y.~Jia, P.~Sermanet, S.~Reed, D.~Anguelov, D.~Erhan,
  V.~Vanhoucke, and A.~Rabinovich, ``Going deeper with convolutions,'' in
  \emph{Proceedings of the IEEE conference on computer vision and pattern
  recognition}, 2015, pp. 1--9.

\bibitem{he2016deep}
K.~He, X.~Zhang, S.~Ren, and J.~Sun, ``Deep residual learning for image
  recognition,'' in \emph{Proceedings of the IEEE conference on computer vision
  and pattern recognition}, 2016, pp. 770--778.

\bibitem{Zhang2018TheUE}
R.~Zhang, P.~Isola, A.~A. Efros, E.~Shechtman, and O.~Wang, ``The unreasonable
  effectiveness of deep features as a perceptual metric,'' \emph{2018 IEEE/CVF
  Conference on Computer Vision and Pattern Recognition}, pp. 586--595, 2018.

\bibitem{Johnson2016PerceptualLF}
J.~Johnson, A.~Alahi, and L.~Fei-Fei, ``Perceptual losses for real-time style
  transfer and super-resolution,'' in \emph{ECCV}, 2016.

\bibitem{Krizhevsky2012ImageNetCW}
A.~Krizhevsky, I.~Sutskever, and G.~E. Hinton, ``Imagenet classification with
  deep convolutional neural networks,'' \emph{Communications of the ACM},
  vol.~60, pp. 84 -- 90, 2012.

\bibitem{Deng2019ArcFaceAA}
J.~Deng, J.~Guo, and S.~Zafeiriou, ``Arcface: Additive angular margin loss for
  deep face recognition,'' \emph{2019 IEEE/CVF Conference on Computer Vision
  and Pattern Recognition (CVPR)}, pp. 4685--4694, 2019.

\bibitem{FaceParsing}
Z.~Liu,
  \url{https://github.com/switchablenorms/CelebAMask-HQ/tree/master/face\_parsing},
  accessed: Mar. 2021. [Online].

\bibitem{Hu2020SqueezeandExcitationN}
J.~Hu, L.~Shen, S.~Albanie, G.~Sun, and E.~Wu, ``Squeeze-and-excitation
  networks,'' \emph{IEEE Transactions on Pattern Analysis and Machine
  Intelligence}, vol.~42, pp. 2011--2023, 2020.

\bibitem{Liu2020OnTV}
L.~Liu, H.~Jiang, P.~He, W.~Chen, X.~Liu, J.~Gao, and J.~Han, ``On the variance
  of the adaptive learning rate and beyond,'' \emph{ArXiv}, vol.
  abs/1908.03265, 2020.

\bibitem{Zhang2019LookaheadOK}
M.~R. Zhang, J.~Lucas, G.~E. Hinton, and J.~Ba, ``Lookahead optimizer: k steps
  forward, 1 step back,'' in \emph{NeurIPS}, 2019.

\bibitem{Hor2010ImageQM}
A.~Hor{\'e} and D.~Ziou, ``Image quality metrics: Psnr vs. ssim,'' \emph{2010
  20th International Conference on Pattern Recognition}, pp. 2366--2369, 2010.

\bibitem{Wang2004ImageQA}
Z.~Wang, A.~Bovik, H.~Sheikh, and E.~P. Simoncelli, ``Image quality assessment:
  from error visibility to structural similarity,'' \emph{IEEE Transactions on
  Image Processing}, vol.~13, pp. 600--612, 2004.

\bibitem{Heusel2017GANsTB}
M.~Heusel, H.~Ramsauer, T.~Unterthiner, B.~Nessler, and S.~Hochreiter, ``Gans
  trained by a two time-scale update rule converge to a local nash
  equilibrium,'' in \emph{NIPS}, 2017.

\bibitem{Huang2020CurricularFaceAC}
Y.~Huang, Y.~Wang, Y.~Tai, X.~Liu, P.~Shen, S.~xin Li, J.~Li, and F.~Huang,
  ``Curricularface: Adaptive curriculum learning loss for deep face
  recognition,'' \emph{2020 IEEE/CVF Conference on Computer Vision and Pattern
  Recognition (CVPR)}, pp. 5900--5909, 2020.

\bibitem{Tov2021DesigningAE}
O.~Tov, Y.~Alaluf, Y.~Nitzan, O.~Patashnik, and D.~Cohen-Or, ``Designing an
  encoder for stylegan image manipulation,'' \emph{ACM Transactions on Graphics
  (TOG)}, vol.~40, pp. 1 -- 14, 2021.

\bibitem{Krause20133DOR}
J.~Krause, M.~Stark, J.~Deng, and L.~Fei-Fei, ``3d object representations for
  fine-grained categorization,'' \emph{2013 IEEE International Conference on
  Computer Vision Workshops}, pp. 554--561, 2013.

\bibitem{yu2015lsun}
F.~Yu, A.~Seff, Y.~Zhang, S.~Song, T.~Funkhouser, and J.~Xiao, ``Lsun:
  Construction of a large-scale image dataset using deep learning with humans
  in the loop,'' \emph{arXiv preprint arXiv:1506.03365}, 2015.

\bibitem{Zhang2020UDHUD}
C.~Zhang, P.~Benz, A.~Karjauv, G.~Sun, and I.-S. Kweon, ``Udh: Universal deep
  hiding for steganography, watermarking, and light field messaging,'' in
  \emph{NeurIPS}, 2020.

\bibitem{Baluja2017HidingII}
S.~Baluja, ``Hiding images in plain sight: Deep steganography,'' in
  \emph{NIPS}, 2017.

\bibitem{DBLP:conf/bmvc/NekrasovS018}
V.~Nekrasov, C.~Shen, and I.~Reid, ``Light-weight refinenet for real-time
  semantic segmentation,'' in \emph{British Machine Vision Conference 2018,
  {BMVC} 2018, Newcastle, UK, September 3-6}, 2018.

\bibitem{Holub2012DesigningSD}
V.~Holub and J.~Fridrich, ``Designing steganographic distortion using
  directional filters,'' \emph{2012 IEEE International Workshop on Information
  Forensics and Security (WIFS)}, pp. 234--239, 2012.

\bibitem{Holub2014UniversalDF}
V.~Holub, J.~Fridrich, and T.~Denemark, ``Universal distortion function for
  steganography in an arbitrary domain,'' \emph{EURASIP Journal on Information
  Security}, vol. 2014, pp. 1--13, 2014.

\bibitem{Sedighi2016ContentAdaptiveSB}
V.~Sedighi, R.~Cogranne, and J.~Fridrich, ``Content-adaptive steganography by
  minimizing statistical detectability,'' \emph{IEEE Transactions on
  Information Forensics and Security}, vol.~11, pp. 221--234, 2016.

\bibitem{zhang2020model}
J.~Zhang, D.~Chen, J.~Liao, H.~Fang, W.~Zhang, W.~Zhou, H.~Cui, and N.~Yu,
  ``Model watermarking for image processing networks,'' in \emph{Proceedings of
  the AAAI Conference on Artificial Intelligence}, vol.~34, no.~07, 2020, pp.
  12\,805--12\,812.

\bibitem{Filler2011MinimizingAD}
T.~Filler, J.~Judas, and J.~Fridrich, ``Minimizing additive distortion in
  steganography using syndrome-trellis codes,'' \emph{IEEE Transactions on
  Information Forensics and Security}, vol.~6, pp. 920--935, 2011.

\bibitem{FaceSwap}
FaceSwap, \url{https://github.com/wuhuikai/FaceSwap}, accessed: Feb. 2021.
  [Online].

\bibitem{Li2019FaceShifterTH}
L.~Li, J.~Bao, H.~Yang, D.~Chen, and F.~Wen, ``Faceshifter: Towards high
  fidelity and occlusion aware face swapping,'' \emph{ArXiv}, vol.
  abs/1912.13457, 2019.

\bibitem{CelebAMask-HQ}
C.-H. Lee, Z.~Liu, L.~Wu, and P.~Luo, ``Maskgan: Towards diverse and
  interactive facial image manipulation,'' in \emph{IEEE Conference on Computer
  Vision and Pattern Recognition (CVPR)}, 2020.

\end{thebibliography}

\end{document}